\documentclass[10pt,twocolumn,letterpaper]{article}

\usepackage{dicta}
\usepackage{times}
\usepackage{epsfig}
\usepackage{graphicx}
\usepackage{amsmath}
\usepackage{amssymb}
\usepackage[T1]{fontenc}
\usepackage[utf8]{inputenc}
\usepackage{longtable}   % multi-page supplementary tables
\usepackage{pdflscape}   % landscape pages for wide supplementary tables

\usepackage[pagebackref=true,breaklinks=true,letterpaper=true,colorlinks,bookmarks=false]{hyperref}

\dictafinalcopy % camera-ready / arXiv: named authors, no line numbers, no review header

\makeatletter
\def\@maketitle{%
   \newpage
   \null
   \vskip -0.35in
   \begin{center}
      {\Large \bf \@title \par}
      \vspace*{24pt}
      {\large \lineskip .5em
       \begin{tabular}[t]{c}\@author\end{tabular}\par}
      \vskip .5em
      \vspace*{12pt}
   \end{center}}
\makeatother

\def\dictaPaperID{****} % *** Enter the DICTA Paper ID here

\begin{document}

%%%%%%%%% TITLE
\title{SAMRI-3D: Adapting SAM2 for 3D MRI Segmentation with Global Volume Tokens}

\author{Zhao Wang$^{1,\ast}$, Wei Dai$^{1}$, Hongfu Sun$^{2}$, Craig Engstrom$^{3}$, and Shekhar S. Chandra$^{1}$\\
{\normalsize $^{1}$School of Electrical Engineering and Computer Science, The University of Queensland, Brisbane, Australia}\\
{\normalsize $^{2}$School of Engineering, College of Engineering, Science and Environment, University of Newcastle, Australia}\\
{\normalsize $^{3}$School of Human Movement and Nutrition Sciences, The University of Queensland, Australia}\\
{\tt\small $^{\ast}$Corresponding author: zhao.wang1@uq.edu.au}
}

\maketitle

%%%%%%%%% ABSTRACT
\begin{abstract}
   Foundation models such as Segment Anything Model 2 (SAM2) have transformed
   natural-image and video segmentation, and recent work has begun adapting them to
   medical imaging. These
   adaptations, however, are largely general-purpose models that treat MRI as one
   modality among many; large-scale, MRI-specific modelling and benchmarking remain
   limited, even though MRI's low soft-tissue contrast leaves many boundaries
   effectively invisible on individual slices.
   We present SAMRI-3D, a benchmark and method for 3D MRI segmentation with
   SAM2. The SAMRI-3D benchmark is the largest MRI-only evaluation to date ---
   10,392 volumes from 34 datasets (27 public, 7 in-house) spanning 12
   anatomical domains and 10+ sequences, with explicit seen/unseen splits.
   Freezing the image encoder and fine-tuning only the lightweight decoder and
   memory modules raises mean Dice from 0.58 (zero-shot SAM2) to 0.76,
   surpassing recent SAM-based medical models (SAMed-2 0.69, Medical-SAM2 0.49,
   SAM-Med3D 0.37) with strong statistical significance. To target invisible
   boundaries, we introduce Global Volume Tokens (GVT): persistent memory tokens
   trained with a Truncated Signed Distance Field (TSDF) reconstruction objective
   that is discarded at inference (zero added cost). This full model, SAMRI-3D,
   attains the best accuracy (0.78) and lowest variance across all 34 datasets and,
   uniquely, shows no drop on 8 held-out datasets (0.79 unseen vs.\ 0.78 seen);
   per-sequence analysis confirms the TSDF objective helps most where per-slice
   contrast is weakest. We will release the benchmark, code, and models in this paper.
\end{abstract}


%%%%%%%%% BODY TEXT
\section{Introduction}

\begin{figure*}[t]
\begin{center}
\includegraphics[width=\textwidth]{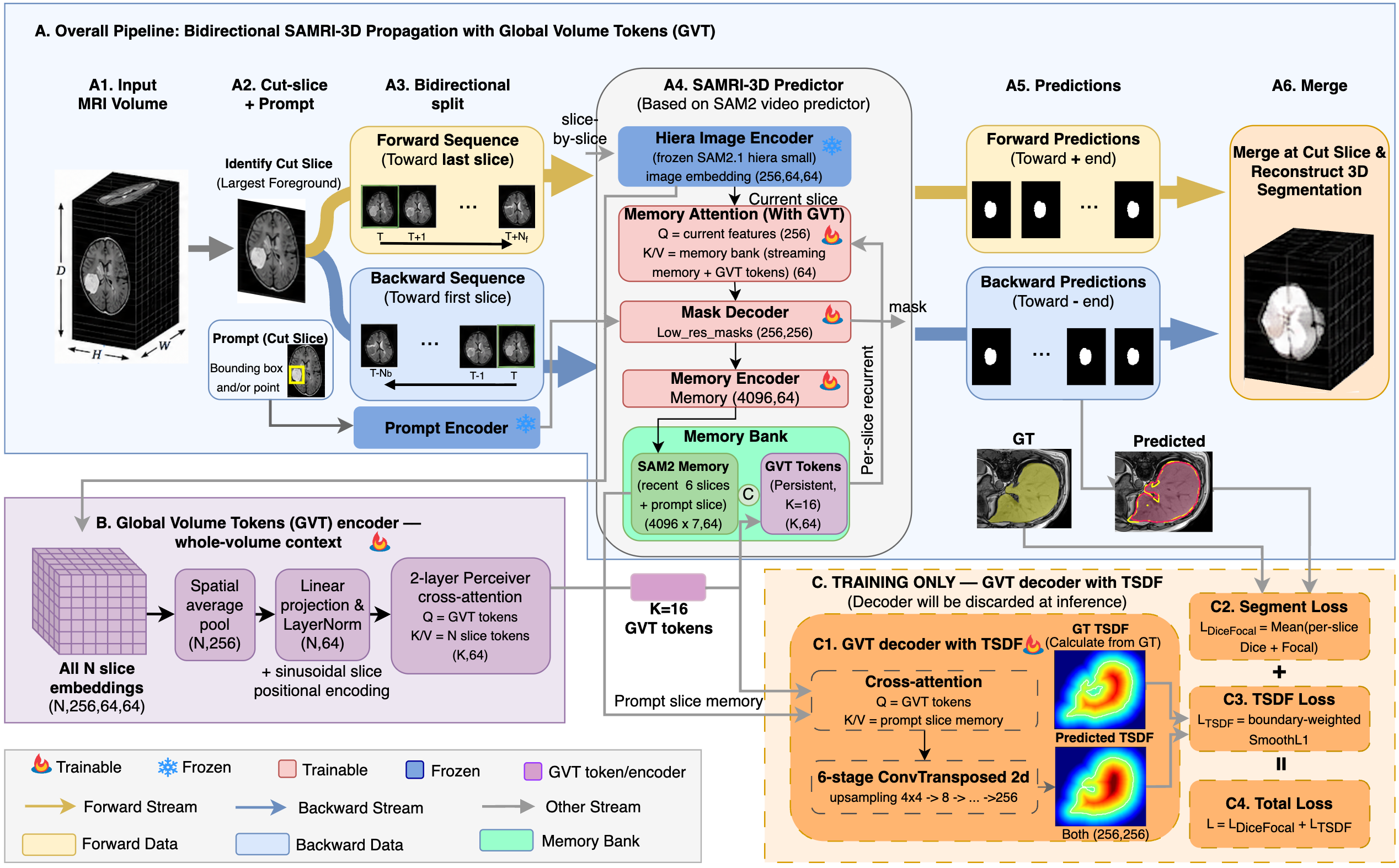}
\end{center}
   \caption{End-to-end SAMRI-3D workflow. \textbf{(A)} Given a 3D MRI volume and a
   box/point prompt on its cut slice (the slice with the largest target
   foreground), the model splits the volume into a forward and a backward
   sub-sequence and segments it slice by slice, propagating outward from the cut
   slice in both directions before merging the two halves into the final 3D mask.
   The predictor is SAM2 with a frozen image encoder and fine-tuned decoder and
   memory modules; at every step its memory attention reads a streaming memory bank
   that we augment with a persistent set of Global Volume Tokens (GVT).
   \textbf{(B)} The GVT encoder builds these tokens by compressing the embeddings of
   all slices into a compact summary of the whole volume, which is injected into the
   memory bank so that every slice---however far from the cut slice---sees
   global context. \textbf{(C)} During training only, a lightweight decoder uses the
   GVT tokens to reconstruct a Truncated Signed Distance Field (TSDF) of the target
   and is supervised against the ground-truth TSDF; this reconstruction objective
   forces the tokens to encode boundary geometry. The network is trained with the
   segmentation loss plus this TSDF loss, and the TSDF decoder is discarded at
   inference, adding no cost.}
\label{fig:pipeline}
\end{figure*}

Magnetic resonance imaging (MRI) segmentation underpins diagnosis, treatment
planning, and disease monitoring, yet it is difficult: intensities are non-uniform,
contrast varies with protocol, and many important structures are small or weakly
delineated. Manual delineation is accurate but costly and varies between experts.
Supervised networks --- whether 2D or 3D, from nnU-Net \cite{ref5} and SwinUNETR
\cite{ref6} to specialised MRI models \cite{ref30,ref31,ref32} --- are accurate but
must be retrained whenever the target or protocol changes. Promptable foundation
models such as the Segment Anything Model (SAM) \cite{ref1} offer an alternative ---
segmenting arbitrary targets from simple prompts without per-target retraining ---
but SAM is trained on natural images and transfers poorly to medical data.

Medical adaptations such as MedSAM \cite{ref3} and the MRI-specific SAMRI
\cite{ref28} reduce this domain gap but remain 2D: applied to a volume they
segment each slice independently, ignoring through-plane continuity. SAM2
\cite{ref2} extends SAM to video, propagating predictions across frames through a
streaming memory bank --- a design that maps naturally onto 3D volumes via the
``slice-as-frame'' paradigm (Section~2.1). Still, SAM2 underperforms out-of-the-box
on medical scans, and its finite memory bank loses information from distant slices
in long volumes.

Existing SAM-for-3D adaptations \cite{ref8,ref9,ref10,ref11,ref23,ref24} instead add
dedicated 3D modules or discard SAM2's streaming memory entirely. We keep that
memory and address two open problems. First, existing SAM2 medical evaluations each
cover only a few organs or a single modality; no large-scale, MRI-specific benchmark
exists. Second, SAM2's memory bank provides only local inter-slice coherence and
lacks global volumetric context --- information from the whole volume that could
guide ambiguous slices; recent memory improvements \cite{ref20,ref7} still operate
within a local temporal window. We propose Global Volume Tokens (GVT) to inject
whole-volume context into SAM2's memory bank, trained with a Truncated Signed
Distance Field (TSDF) objective that encodes boundary geometry rather than redundant
binary masks.

Our contributions are: (1) the SAMRI-3D benchmark, the largest MRI-only SAM2
evaluation to date --- 10,392 volumes from 34 datasets (27 public, 7 in-house),
12 anatomical domains, and 10+ sequences, with seen/unseen splits for zero-shot
testing;
(2) an efficient MRI adaptation of SAM2 that freezes the image encoder and
fine-tunes only the lightweight decoder and memory modules, lifting mean Dice
from 0.58 (SAM2 zero-shot) to 0.76 and already outperforming recent SAM-based
medical models by statistically significant margins; (3) Global Volume
Tokens (GVT) with a Truncated Signed Distance Field (TSDF) reconstruction
objective --- the first study of compact, persistent whole-volume memory tokens for
SAM2; we compare training these tokens with segmentation loss alone, auxiliary
binary reconstruction, and TSDF reconstruction, and show the TSDF variant is the
most effective, giving the best accuracy (0.78) and lowest variance at zero
inference overhead; and (4) robust generalisation and mechanistic analysis --- our
model transfers to held-out datasets with no performance drop, and per-sequence
analysis shows the reconstruction objective helps exactly where image boundaries
are weakest.

\section{Related Work}

\subsection{SAM2 and Its Use for 3D MRI}

SAM2 \cite{ref2} is a promptable segmentation model for both images and video,
pairing a Hiera image encoder, a prompt encoder (points or boxes), and a mask
decoder with a streaming memory bank. A video is segmented frame by frame: when
predicting the current frame, a memory attention module conditions it on features
stored from previously processed frames. Because adjacent frames are highly
correlated, this lets a mask prompted on a few frames propagate coherently through
the whole sequence. The same property holds along the third axis of a 3D volume,
where neighbouring slices are strongly correlated, so SAM2 fits 3D MRI under the
slice-as-frame paradigm: the volume is treated as a video and a mask prompted on one
slice is propagated slice by slice, with the memory bank carrying the inter-slice
context that a purely 2D model would discard \cite{ref18,ref19}. This through-plane
coherence is precisely why we build on SAM2 rather than a 2D segmenter; we detail
the memory bank, and its limits, in Section~2.2. Medical adaptations of SAM2
fine-tune it or adjust its memory --- MedSAM2 \cite{ref4} on 455K+ image-mask pairs,
SAMed-2 \cite{ref7} with temporal adapters and confidence-driven memory, and RevSAM2
\cite{ref21} with reverse propagation --- but are evaluated predominantly on CT or
limited MRI subsets.

\subsection{Memory Bank and Global Context}

The memory bank is the component we build on. A memory encoder stores each processed
frame as a spatial feature map paired with its predicted mask, and the memory
attention retrieves a fixed number of the most recent such entries when segmenting
the current frame. Because SAM2 is built to segment unbounded video, this makes the
bank a sliding window: it gives strong local coherence between neighbouring slices
but evicts distant ones, so information from far parts of a long volume is lost. An MRI volume, however, is
finite --- at most a couple of hundred slices --- so its entire content can be
summarised once and made globally available, which is exactly what our Global
Volume Tokens do (Section~3.3). Prior work instead improves memory quality within
the local window --- short/long-term banks (SLM-SAM2 \cite{ref20}), self-sorting
confidence-based selection (MedSAM-2 \cite{ref22}), and confidence-driven
replacement (SAMed-2 \cite{ref7}) --- but none provides whole-volume context.
Compressing global context into a compact set of learnable tokens has precedent in
MRI: Bran Lorenzana \etal \cite{ref29} used global image tokens for
transformer-based reconstruction.

\subsection{Signed Distance Fields}

A signed distance field (SDF) replaces a binary mask with each pixel's signed
distance to the nearest boundary --- positive inside, negative outside, zero on the
boundary --- a dense, geometry-aware target that emphasises boundary localisation
over per-pixel occupancy. Recent segmentation work exploits this through
boundary-weighted losses (FocusSDF \cite{ref25}), flow-matching (FlowSDF
\cite{ref26}), and SDF-supervised token compression (TokenSeg \cite{ref27}). A full
SDF is unbounded, so much of its loss is spent on voxels far from the boundary; a
truncated SDF (TSDF) clamps the field to a narrow band around the boundary,
concentrating supervision where it matters. We compute a per-slice TSDF from the
ground-truth mask (see supplementary) and use it as the reconstruction target for
our Global Volume Tokens (Section~3.3).

\section{Method}

\subsection{Problem Formulation}

Given a 3D MRI volume $V \in \mathbb{R}^{D\times H\times W}$, we treat it as a
$D$-frame sequence and identify the \emph{cut slice} \cite{ref19} --- the slice with
the largest target foreground --- which gives the clearest view of the target and
the most confident single-slice prediction, the best anchor for propagation. We
split the volume at this slice into a forward and a backward sub-sequence
(Fig.~\ref{fig:pipeline}(A)), each processed independently by SAMRI-3D's video
predictor (fine-tuned from SAM2) with the cut slice as the prompted frame. During
training the prompt is a bounding box derived from the ground-truth mask on the cut
slice; at inference, boxes and/or points are supplied by the user or an upstream
detector. Predictions propagate outward from the anchor in both directions and are
merged at the cut slice into the full 3D mask, preserving the original NIfTI spatial
metadata.

\subsection{Architecture}

Following the parameter-efficient strategy of our prior 2D adaptation SAMRI
\cite{ref28}, we freeze the Hiera image encoder and the prompt encoder and
fine-tune only the lightweight decoder and memory modules (mask decoder, memory
encoder, and memory attention). Because the encoder is frozen, its embeddings are
pre-computed once for the whole training set ($\sim$3.3\,TB), avoiding repeated
encoder passes and greatly reducing training cost. The model then operates on three
complementary levels of spatial context: within-slice features from the frozen
encoder; local inter-slice coherence from the streaming memory bank; and
whole-volume context from the Global Volume Tokens (Section~3.3), which persist in
the memory bank so that every slice, however far from the cut slice, sees a global
summary.

\subsection{Global Volume Tokens}

To give every slice whole-volume context, we introduce Global Volume Tokens (GVT).
The GVT encoder (Fig.~\ref{fig:pipeline}(B)) takes the pre-computed Hiera features
of all $D$ slices, spatially average-pools each into a slice vector, adds a
sinusoidal positional encoding along the volume axis, and compresses the sequence
with a 2-layer Perceiver-style \cite{ref13} cross-attention into $K=16$ persistent
tokens (dimension $C_\mathrm{mem}=64$). These tokens are appended to the memory bank
at every propagation step, so --- unlike the sliding-window memory --- they remain
visible however far a slice is from the cut slice.

Trained through the segmentation loss from SAM2 alone, however, the tokens tend to become
redundant with the memory bank: the memory attention's softmax competition lets the
per-frame entries dominate, leaving the tokens little direct signal. We therefore
supervise them with an auxiliary reconstruction objective (Fig.~\ref{fig:pipeline}(C)).
A training-only decoder cross-attends the GVT tokens ($Q$) to the prompted-slice
memory features ($K/V$) --- a natural label identifier, so the same architecture
generalises across all structures without a label-embedding table --- and upsamples
them, through 6-stage ConvTranspose2d from $4\times4$ to $256\times256$, into a
Truncated Signed Distance Field (TSDF) of the target. This is compared to the
ground-truth TSDF (truncation radius $\tau=32$ px) by a boundary-weighted SmoothL1
loss that upweights voxels near the boundary shell. Because a TSDF encodes boundary
geometry at every voxel rather than binary occupancy, it forces the tokens to carry
shape information complementary to the appearance-based memory --- unlike a binary
reconstruction target, which duplicates what the mask decoder already produces. The
decoder is discarded at inference, so GVT adds no test-time cost; the supplementary
gives full layer configurations, tensor shapes, and the trainable/frozen breakdown.
The model is trained with
\begin{equation}
\mathcal{L} = \mathcal{L}_\mathrm{DiceFocal} + \mathcal{L}_\mathrm{TSDF}.
\end{equation}

\section{Experiments and Results}

\subsection{Dataset}

The SAMRI-3D benchmark comprises 10,392 3D MRI volumes from 34 datasets (27
public, 7 in-house) spanning 12 anatomical domains. By comparison, prior SAM2
medical studies are evaluated on far fewer datasets --- for example, a single
knee-MRI dataset \cite{ref18}, or BraTS plus four Medical Segmentation Decathlon
tasks \cite{ref19} --- so no existing study assesses SAM2 on MRI at this scale. We
partition the 34 datasets into 26 seen and 8 unseen datasets; each dataset is
split 80/10/10 into train/validation/test. Models are trained and validated only
on the seen datasets' train/validation subsets and evaluated on the test subsets
of both the seen and unseen datasets, the latter measuring zero-shot
generalisation to datasets unseen during training. Table~\ref{tab:benchmark}
summarises the benchmark by anatomical domain; the complete list of all 34
datasets with per-dataset sequence, volume count, and seen/unseen split is
provided in Table~S5 of the supplementary material.

\begin{table}[t]
\begin{center}
\footnotesize
\setlength{\tabcolsep}{3pt}
\begin{tabular}{|l|c|r|l|}
\hline
Domain & \# DS & \# Vol. & Sequences \\
\hline\hline
Brain & 8 & 5,520 & FLAIR, T1, T1CE, ADC, DWI, T2 \\
Prostate & 7 & 2,234 & ADC, T2 \\
Knee/MSK & 8 & 1,053 & FLASH, PD, T2, DESS, TSE \\
Cardiac & 3 & 340 & Cine, LGE \\
Whole-body & 1 & 263 & Mixed \\
Hippocampus & 1 & 260 & T1 \\
Vestibular & 1 & 227 & T1CE \\
Hip & 1 & 211 & T2 \\
Spine & 1 & 172 & T2 \\
Abdominal & 1 & 60 & T1 \\
Shoulder & 1 & 45 & T2 \\
Gynecology & 1 & 7 & T2 \\
\hline
Total & 34 & 10,392 & 10+ MRI sequences \\
\hline
\end{tabular}
\end{center}
\caption{SAMRI-3D Benchmark Summary by Anatomical Domain.}
\label{tab:benchmark}
\end{table}

\subsection{Implementation Details}

All models use the SAM2.1 Hiera-Small backbone with pre-computed slice embeddings,
trained from these embeddings with AdamW \cite{ref17} (learning rate
$5\times10^{-6}$, weight decay 0.01). Each volume is trained with an 80\%/20\% mix
of volumetric sequences and 2D slice batches --- the 2D batches preserve strong
per-slice segmentation, which supplies the reliable prompted-slice predictions that
condition subsequent propagation.

To study how the GVT tokens should be trained, we compare four models
(Fig.~S1 in the supplementary): a 2D-slice baseline without GVT (GVT-baseline); GVT tokens
trained through the segmentation loss alone (GVT-v1); with an auxiliary
low-resolution binary reconstruction (GVT-v2); and with the TSDF reconstruction
(GVT-v3, our SAMRI-3D). All use the per-slice Dice + Focal loss \cite{ref16}; GVT-v3
adds the TSDF term of Eq.~(1).

We report mean 3D Dice ($\pm$ std across the 34 datasets), Hausdorff Distance (HD),
and Mean Surface Distance (MSD); per dataset we take the median across volumes, then
the mean and std of these per-dataset medians. Significance uses two-sided paired
Wilcoxon signed-rank tests at the per-sample ($n\approx3{,}376$), per-dataset
($n=34$), and per-label ($n=49$) levels; full statistics in Table~S12.

\subsection{Main Results}

Table~\ref{tab:main} reports the headline comparison: three prior SAM-based medical
methods, zero-shot SAM2-Small, and our SAMRI-3D. Because SAMRI-3D is built on the
Hiera-Small backbone, SAM2-Small is its backbone-matched zero-shot reference.
SAMRI-3D attains the best mean accuracy and lowest variance on every metric, lifting
mean Dice from 0.58 (zero-shot SAM2-Small) to 0.78 and clearly surpassing SAMed-2,
Medical-SAM2, and SAM-Med3D. All improvements over the comparison methods and
SAM2-Small are significant at the per-sample and per-label levels (full statistics
in Table~S12). Two further SAM2-based adaptations produced near-zero Dice on this
benchmark and are omitted. Backbone size has little effect (Section~4.4) and the
token-training objective is analysed in Section~4.5. Per-dataset and per-target
breakdowns are in Tables~S6--S11.

\begin{table}[t]
\begin{center}
\scriptsize
\setlength{\tabcolsep}{3.5pt}
\begin{tabular}{|l|c|c|c|}
\hline
Method & Dice $\uparrow$ & HD $\downarrow$ & MSD $\downarrow$ \\
\hline\hline
SAM-Med3D \cite{ref11} & $0.37\pm0.21$ & $29.04\pm16.70$ & $6.39\pm2.78$ \\
Medical-SAM2 \cite{ref22} & $0.49\pm0.22$ & $33.51\pm21.81$ & $5.92\pm4.26$ \\
SAMed-2 \cite{ref7} & $0.69\pm0.15$ & $22.28\pm18.82$ & $2.91\pm1.62$ \\
SAM2-Small & $0.58\pm0.24$ & $28.24\pm20.42$ & $6.32\pm4.46$ \\
SAMRI-3D (Ours) & $\mathbf{0.78\pm0.14}$ & $\mathbf{13.85\pm9.01}$ & $\mathbf{2.11\pm0.96}$ \\
\hline
\end{tabular}
\end{center}
\caption{Main comparison: three prior SAM-based medical methods, the
backbone-matched zero-shot SAM2-Small, and SAMRI-3D (mean $\pm$ std across 34
datasets). SAM2 backbone sizes and GVT ablations are in Tables~\ref{tab:scale}
and~\ref{tab:ablation}.}
\label{tab:main}
\end{table}

\subsection{SAM2 Scale Analysis}

The aim of this analysis is to justify our backbone choice. SAM2's encoder scale
has little effect on zero-shot MRI accuracy: across the four SAM2 sizes
(Table~\ref{tab:scale}), a $5.8\times$ increase in parameters (38.9M to 224.4M)
brings no improvement --- all four variants fall within a narrow 0.56--0.58 mean
Dice band, and the two largest models are in fact marginally worse, indicating
that the extra capacity trained on natural video transfers poorly to MRI. We
therefore avoid a large backbone and fine-tune from SAM2-Small, whose zero-shot
accuracy is the best (0.58) at near-minimal parameter cost (46.0M, versus 224.4M
for Large), giving the best accuracy--efficiency trade-off for fine-tuning. This
size-insensitivity also supports our frozen-encoder design: since a larger encoder
does not help, we freeze the Hiera encoder and adapt only the decoder and memory
modules.

\begin{table}[t]
\begin{center}
\footnotesize
\setlength{\tabcolsep}{4pt}
\begin{tabular}{|l|r|c|c|c|}
\hline
Backbone & Params & Dice $\uparrow$ & HD $\downarrow$ & MSD $\downarrow$ \\
\hline\hline
SAM2-Tiny & 38.9M & $0.57\pm0.23$ & $27.79\pm20.19$ & $6.73\pm5.07$ \\
SAM2-Small & 46.0M & $\mathbf{0.58\pm0.24}$ & $28.24\pm20.42$ & $6.32\pm4.46$ \\
SAM2-Base+ & 80.8M & $0.56\pm0.24$ & $29.95\pm21.39$ & $6.23\pm4.22$ \\
SAM2-Large & 224.4M & $0.56\pm0.24$ & $30.22\pm20.87$ & $7.83\pm5.43$ \\
\hline
\end{tabular}
\end{center}
\caption{Zero-shot SAM2 across backbone sizes. Accuracy is flat despite a
$5.8\times$ parameter increase, so we fine-tune from SAM2-Small.}
\label{tab:scale}
\end{table}

\subsection{GVT Ablation Analysis}

Table~\ref{tab:ablation} compares the four training strategies. Adding GVT tokens
without a reconstruction objective (GVT-v1) or with a binary reconstruction
(GVT-v2) gives little or only marginal gain over the baseline, whereas the TSDF
reconstruction (GVT-v3, our SAMRI-3D) improves every metric (Dice $0.78\pm0.14$, HD
$13.85\pm9.01$, MSD $2.11\pm0.96$) and yields the lowest Dice variance ($\pm0.14$
vs.\ $\pm0.17$ for the baseline). The gain over the strong baseline is small but
significant on Dice at the per-sample ($p=9.8\times10^{-14}$) and per-label
($p=4.1\times10^{-3}$, 71\% win-rate) levels, though not at the coarser per-dataset
level ($p=0.22$); GVT-v3 also exceeds GVT-v1 at all levels ($p\leq1.1\times10^{-2}$)
while being statistically indistinguishable from GVT-v2, its edge lying in mean
accuracy and reduced variance. This progression is exactly what the design in
Section~3.3 predicts: tokens trained by the segmentation loss alone stay redundant
with the memory bank, a binary reconstruction target duplicates what the mask
decoder already learns, and only the TSDF's boundary geometry supplies information
genuinely complementary to SAM2's appearance-based memory.

\begin{table}[t]
\begin{center}
\scriptsize
\setlength{\tabcolsep}{2pt}
\begin{tabular}{|l|l|c|c|c|}
\hline
Method & Objective & Dice $\uparrow$ & HD $\downarrow$ & MSD $\downarrow$ \\
\hline\hline
GVT-baseline & --- & $0.76\pm0.17$ & $14.21\pm12.32$ & $2.39\pm1.70$ \\
GVT-v1 & seg & $0.76\pm0.16$ & $14.46\pm10.09$ & $2.33\pm1.15$ \\
GVT-v2 & binary & $0.77\pm0.16$ & $13.94\pm10.34$ & $2.25\pm1.40$ \\
GVT-v3 (SAMRI-3D) & TSDF & $\mathbf{0.78\pm0.14}$ & $\mathbf{13.85\pm9.01}$ & $\mathbf{2.11\pm0.96}$ \\
\hline
\end{tabular}
\end{center}
\caption{GVT ablation: how the global tokens are trained, each objective additive
on the per-slice segmentation loss. Objective --- seg: segmentation loss only;
binary: $+$ binary-mask reconstruction; TSDF: $+$ TSDF reconstruction.}
\label{tab:ablation}
\end{table}

\subsection{Per-Domain Highlights}

\begin{figure}[t]
\begin{center}
\includegraphics[width=0.82\linewidth]{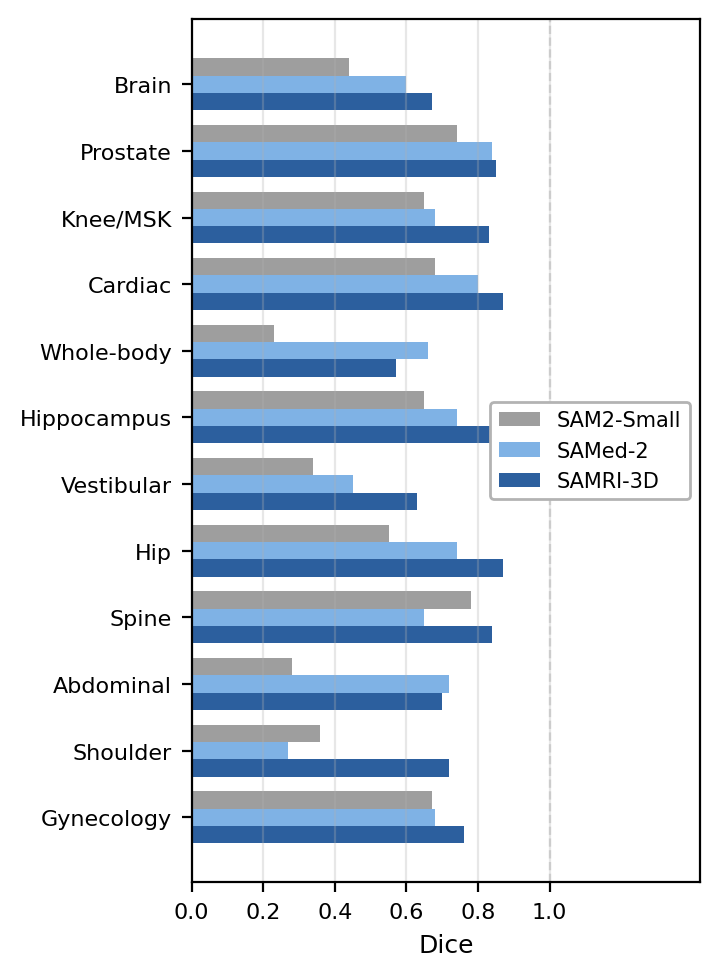}
\end{center}
   \caption{Per-domain Dice (mean of per-dataset medians) for SAMRI-3D versus
   zero-shot SAM2-Small and the strongest external method (SAMed-2) across the 12
   benchmark domains. SAMRI-3D leads in 10 of 12, trailing SAMed-2 only on the
   whole-body and abdominal multi-organ cases.}
\label{fig:perdomain}
\end{figure}

SAMRI-3D leads in 10 of 12 anatomical domains, trailing SAMed-2 only on the
whole-body and abdominal multi-organ cases (Fig.~\ref{fig:perdomain}, Table~S3). It
is strongest on well-defined structures and, more importantly, rescues low-contrast
targets that zero-shot SAM2 largely misses --- white-matter hyperintensities improve
from 0.09 to 0.54 and the hippocampus from 0.65 to 0.84 --- while a few small,
diffuse structures (scapula cartilage, non-enhancing tumour) remain hard for every
method.

\subsection{Per-MRI-Sequence Analysis}

\begin{figure}[t]
\begin{center}
\includegraphics[width=\linewidth]{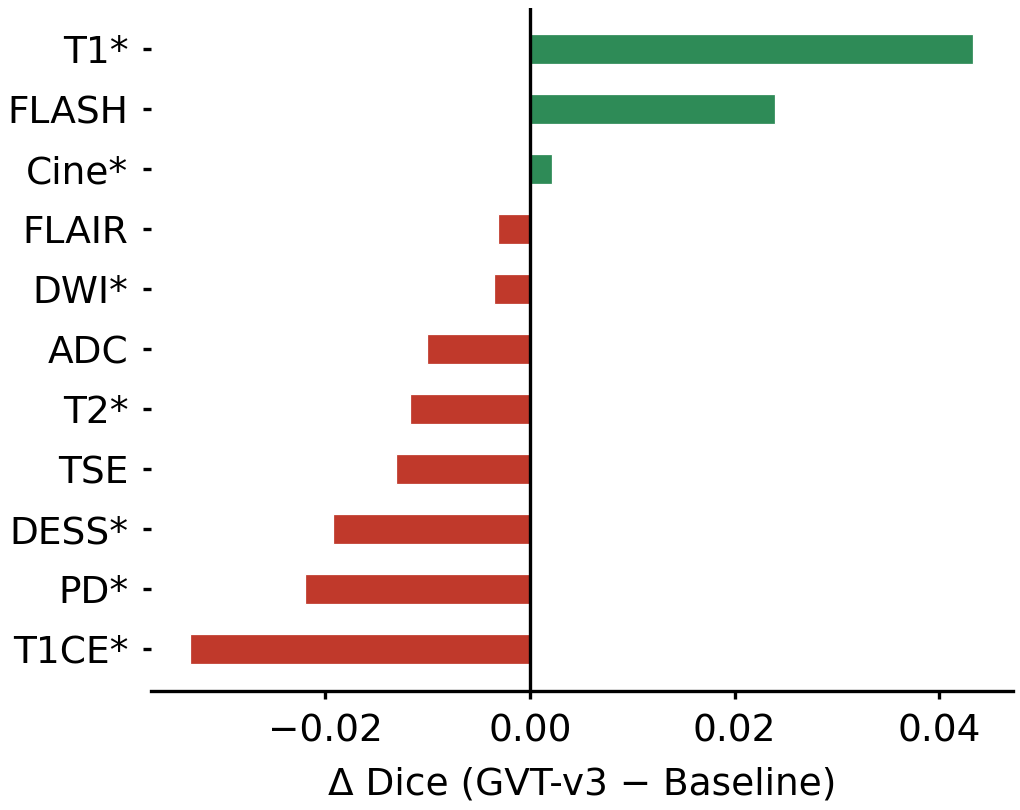}
\end{center}
   \caption{Per-MRI-sequence improvement of SAMRI-3D over the baseline
   (median Dice difference). Green: improvement; red: regression; $*$ marks
   $p<0.05$ (paired Wilcoxon). Single-sequence subset (2,156 samples).}
\label{fig:sequence}
\end{figure}

Because its TSDF objective supplies boundary geometry, SAMRI-3D should help most
where image evidence is weakest, so its effect should track imaging contrast.
Fig.~\ref{fig:sequence} confirms this: across 11 single-sequence categories, gains
concentrate on low-contrast sequences with diffuse boundaries (T1, FLASH; pooled
$+0.035$ mean Dice, $p=1.4\times10^{-6}$) and are neutral-to-slightly-negative on
high-contrast sequences whose boundaries are already sharp (DESS, T1CE). This
sequence dependence is direct evidence for the invisible-boundary motivation behind
the TSDF objective. Full per-sequence values are reported in Table~S4.

\subsection{Zero-Shot Generalisation to Unseen Datasets}

\begin{figure}[t]
\begin{center}
\includegraphics[width=\linewidth]{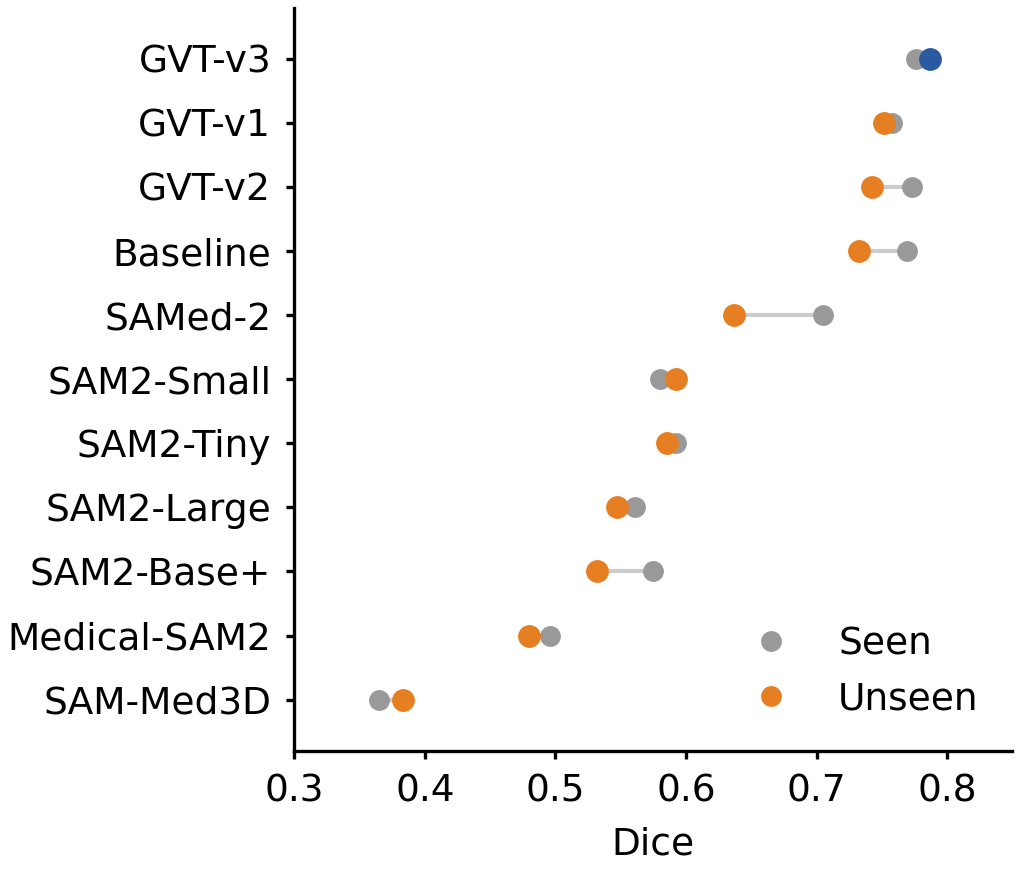}
\end{center}
   \caption{Zero-shot generalisation: per-model seen vs.\ unseen Dice (mean across
   datasets). SAMRI-3D (highlighted) shows no generalisation gap. Seen: 26
   datasets used for training/validation; unseen: 8 datasets excluded from training.}
\label{fig:generalisation}
\end{figure}

Because the 8 unseen datasets contribute no training data, their test subsets probe
generalisation to datasets absent from training. Fig.~\ref{fig:generalisation}
compares seen- and unseen-dataset performance for all models. SAMRI-3D attains the
highest unseen Dice (0.79) and is the only strong model with no generalisation gap
($+0.01$ relative to seen data), whereas SAMed-2 and the fine-tuned baseline both
degrade on unseen data. On the unseen datasets it significantly exceeds SAMed-2 and
zero-shot SAM2 ($p<0.02$); its edge over the fine-tuned baseline is positive but not
significant at $n=8$. Gains on unseen musculoskeletal data (e.g., MSK-FLASH,
$0.38\rightarrow0.71$) show the TSDF objective transfers robustly.

\begin{figure*}[p]
\begin{center}
\includegraphics[width=0.82\textwidth]{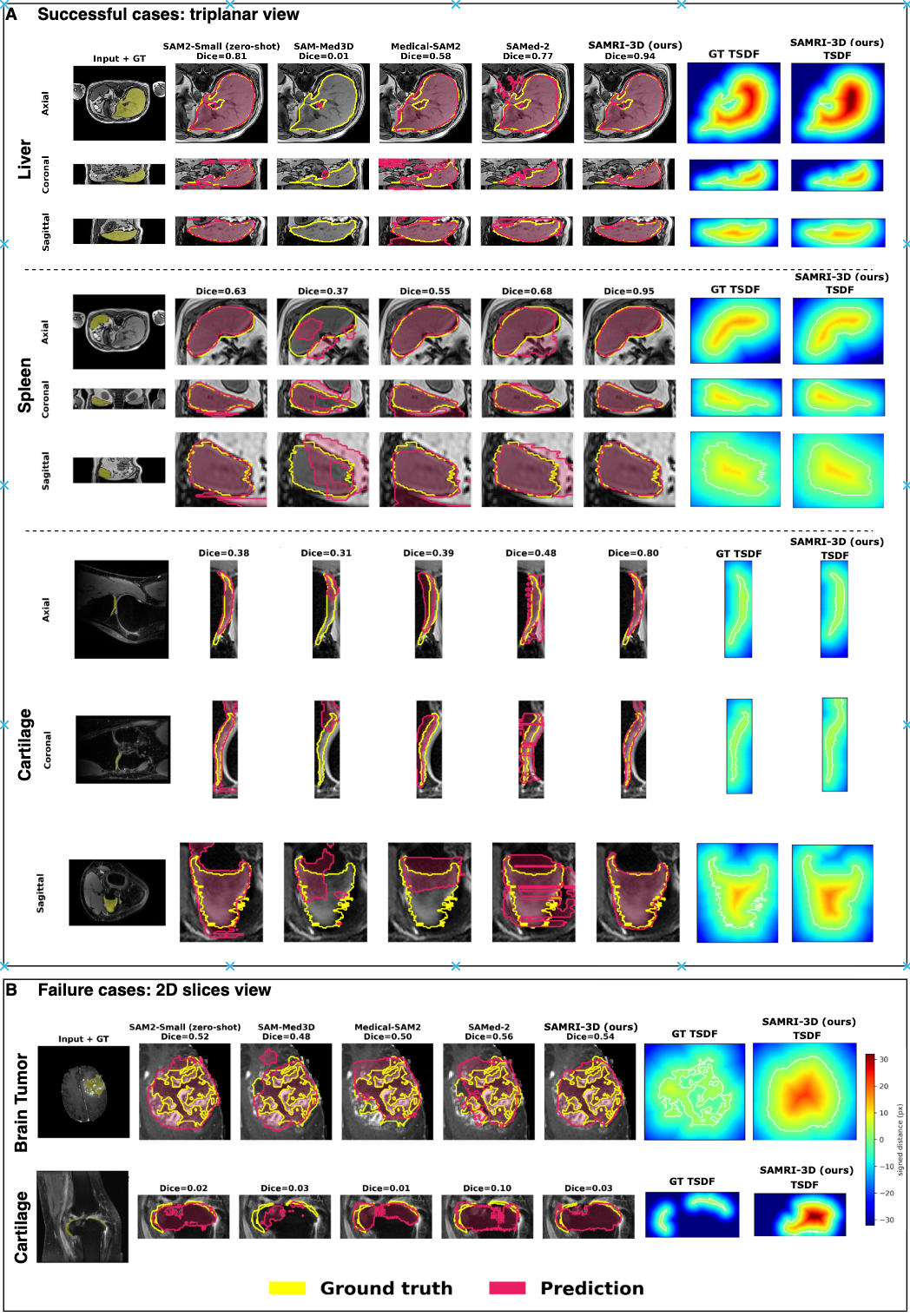}
\end{center}
   \caption{Qualitative segmentation results. Top rows A: representative successful
   cases shown in triplanar (axial, coronal, sagittal) views, with ground-truth and
   predicted masks overlaid; the two rightmost columns show the target's ground-truth
   TSDF and the TSDF reconstructed by SAMRI-3D's training-only decoder. Bottom rows B:
   two characteristic failure modes ---
   (i) a brain tumour whose ground truth is highly fragmented, which almost all
   models instead predict as a single contiguous region; and (ii) a thin cartilage
   whose bounding-box prompt spans a large area, producing an over-segmentation on
   the prompted slice that then propagates into an incorrect 3D volume.}
\label{fig:qualitative}
\end{figure*}

\subsection{Qualitative Results}

Fig.~\ref{fig:qualitative} compares SAMRI-3D against zero-shot SAM2 (Small),
SAM-Med3D, Medical-SAM2, and SAMed-2. Successful cases are shown in triplanar
(axial, coronal, sagittal) views for liver, spleen, and cartilage; SAMRI-3D produces
the most complete and boundary-accurate masks, recovering structure that the
zero-shot and external models under-segment or miss.
The failure cases, shown as single slices, illustrate that all methods, including
SAMRI-3D, struggle on extremely small or diffuse targets, such as a scattered brain
tumour and a thin cartilage, where every model attains low and closely clustered
Dice, reflecting genuine task difficulty rather than a model-specific weakness.
Additional qualitative examples across more anatomical domains are provided in
Fig.~S2 of the supplementary material.

\section{Discussion}

Frozen-encoder fine-tuning works because of the large domain gap between natural
video and MRI: the frozen Hiera encoder supplies general visual features while the
trained decoder and memory adapt them to medical imaging, preserving the encoder's
generalisation.

The finding that model size does not matter for zero-shot medical performance is
notable. All SAM2 variants, from Tiny to Large, achieve similarly low zero-shot
accuracy, suggesting that the bottleneck is not how well the model ``sees'' the
image (the frozen encoder already extracts strong general features) but how well
the decoder and memory interpret those features for the medical domain. Among
existing SAM-based medical methods, SAMed-2's relatively strong performance confirms
the value of adapter-based fine-tuning.

The GVT progression also clarifies how global context should enter SAM2's memory:
only the TSDF reconstruction, which encodes boundary geometry rather than
appearance, adds information that survives the memory attention's softmax
competition (Section 4.5). Its improvement is modest but consistent across all
metrics and datasets, with notably reduced variance; as the significance tests show,
the gain is robust per-sample and per-label but not per-dataset, so we position
SAMRI-3D as the lowest-variance, best-mean model rather than claiming a large
dataset-level improvement.

Several directions remain open. Gated attention that bypasses softmax competition
could amplify SAMRI-3D's gains by modulating features directly rather than competing
for attention weight, and adapting the frozen encoder may lift its performance
ceiling. Small-lesion segmentation is still hard (e.g., adrenal Dice $\leq0.10$),
and comparison with fully-supervised methods such as nnU-Net would further
contextualise the results. Finally, the two failure modes in
Fig.~\ref{fig:qualitative} --- fragmented ground truth collapsed into one contiguous
mask, and over-segmentation from oversized box prompts propagating through the
memory bank --- motivate scatter-aware prompting and tighter prompt-region handling.

\section{Conclusion}

We present SAMRI-3D, a comprehensive benchmark and method for adapting SAM2 to 3D
MRI segmentation. Our best model, SAMRI-3D with TSDF reconstruction, achieves the
best mean accuracy with the lowest variance across 34 diverse datasets,
substantially outperforming three existing SAM-based medical segmentation methods.
We show that TSDF-guided Global Volume Tokens provide consistent improvements where
simpler token formulations fall short, pointing toward structured
geometric priors as a principled approach for global context integration. The
SAMRI-3D benchmark, comprising 10,392 volumes across 12 anatomical domains, provides
a standardised evaluation framework for future MRI segmentation research.

\section*{Code Availability}
Code and pre-trained models are publicly available at
\url{https://github.com/wangzhaomxy/SAMRI-3D}.

\section*{Acknowledgements}
This work was partially supported by a National Health and Medical Research Council
(NHMRC) Ideas Grant (Award No.\ APP2001734). During manuscript preparation the
authors used a large language model to check grammar and refine language; all such
content was reviewed and edited by the authors, who take full responsibility for it.

%%%%%%%%% REFERENCES

\clearpage
\onecolumn
\setcounter{secnumdepth}{0}
\section*{\centering Supplementary Material}

\subsection*{A.\quad TSDF Computation}
For each slice we convert the ground-truth binary mask into a Truncated Signed
Distance Field (TSDF). Let $d_{\mathrm{in}}$ and $d_{\mathrm{out}}$ be the Euclidean
distance transforms of the foreground and background. The signed distance field
\begin{equation}
\mathrm{SDF} = d_{\mathrm{in}} - d_{\mathrm{out}}
\end{equation}
is positive inside the structure, negative outside, and zero on the boundary. We
truncate it to a narrow band,
\begin{equation}
\mathrm{TSDF} = \operatorname{clip}(\mathrm{SDF},\, -\tau,\, +\tau), \qquad \tau = 32~\text{px},
\end{equation}
so that an all-background slice maps to the constant $-\tau$; targets are computed at
$256\times256$ resolution. The TSDF decoder is supervised with a boundary-weighted
Smooth-L1 loss
\begin{equation}
\mathcal{L}_{\mathrm{TSDF}} = \operatorname{mean}\!\big(w \odot \operatorname{SmoothL1}(\hat{y},\, y)\big),
\qquad w = 1 + \alpha\,\exp\!\Big(\!-\tfrac{y^{2}}{2\sigma^{2}}\Big),
\end{equation}
where the weight $w$ up-weights voxels near the boundary shell ($|y|\approx 0$), with
$\alpha=4$ and $\sigma=\tau/3\approx10.7$. For multi-label volumes the per-label loss is
divided by $\sqrt{\mathrm{area}}$ to balance small and large structures.

\subsection*{B.\quad Architecture and Implementation Details}
This section specifies the SAMRI-3D components and their tensor shapes for
reproducibility; all values are for the SAM2.1 Hiera-Small backbone. Below, $N$ is the
number of slices in a volume, $K$ the number of Global Volume Tokens, $C_{\mathrm{mem}}$
the SAM2 memory dimension, and $C_{\mathrm{hid}}$ the encoder/hidden dimension.

\noindent\textbf{Key dimensions.} The constants used throughout are summarised below.

\begin{center}{\footnotesize\setlength{\tabcolsep}{4pt}
\begin{tabular}{|l|l|c|}
\hline
\textbf{Symbol} & \textbf{Meaning} & \textbf{Value} \\ \hline\hline
N & slices per volume & ~50--200 \\
K & Global Volume Tokens & 16 \\
C\_mem & SAM2 memory dimension & 64 \\
C\_hid & encoder / hidden dimension & 256 \\
--- & input resolution & 1024 $\times$ 1024 \\
--- & encoder feature map & 64 $\times$ 64 \\
M & memory tokens per frame (64$\times$64) & 4096 \\
--- & TSDF resolution & 256 $\times$ 256 \\
$\tau$ & TSDF truncation radius & 32 px \\
num\_maskmem & memory sliding window & $\approx$ 7 frames \\
max\_slices & per-video slice cap (training) & 32 \\
\hline
\end{tabular}}\end{center}

\noindent\textbf{GVT encoder.} The encoder compresses all $N$ pre-computed Hiera slice embeddings into $K$ global tokens, run once per volume. Each slice embedding ($256\times64\times64$) is spatially average-pooled to a $256$-d vector; a linear projection maps it to $64$-d and a fixed sinusoidal positional encoding along the slice axis is added, giving an $(N,64)$ sequence. $K=16$ learnable query tokens then cross-attend to this sequence through a 2-layer Perceiver-style Transformer decoder (4 heads, feed-forward width 256), producing $K$ global tokens of dimension 64. The encoder has ${\approx}120$K parameters.
\begin{center}
\includegraphics[width=0.72\linewidth]{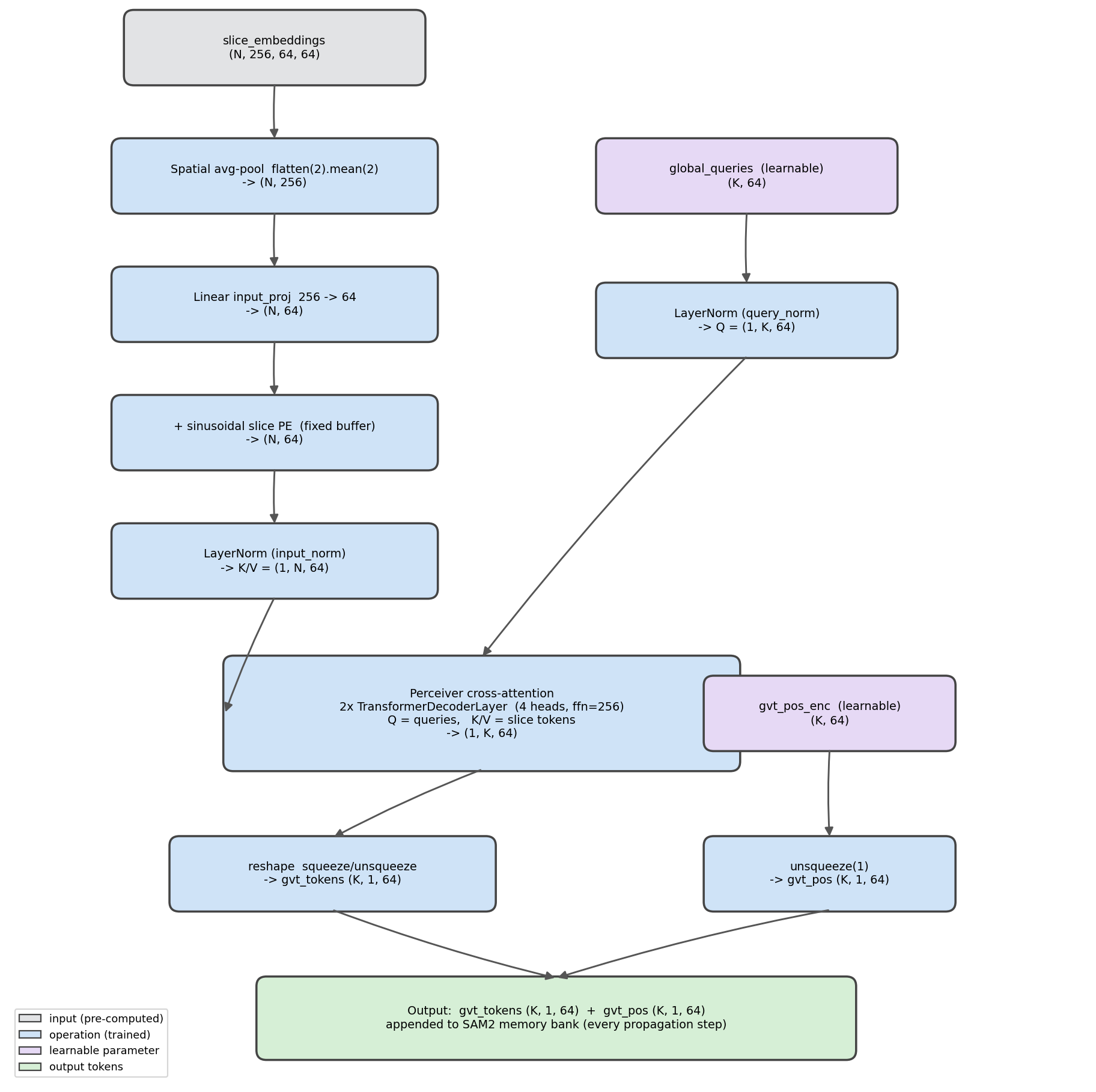}\\[3pt]
{\footnotesize\itshape GVT encoder data flow with tensor shapes.}
\end{center}

\noindent\textbf{Memory-bank fusion.} SAM2s memory attention is wrapped so that, at every propagation step, the cached GVT tokens are concatenated onto the memory bank: the memory sequence grows from $(M_{\mathrm{total}},B,64)$ to $(M_{\mathrm{total}}{+}K,B,64)$. Here $M_{\mathrm{total}}$ is SAM2s sliding window of ${\approx}7$ recent frames (each $64{\times}64{=}4096$ tokens) plus object-pointer tokens; the GVT tokens are treated like object-pointer tokens and excluded from spatial rotary position encoding. The current-frame query and attention output are $C_{\mathrm{hid}}{=}256$-dimensional, whereas the memory keys/values (and the GVT tokens) are $C_{\mathrm{mem}}{=}64$-dimensional; cross-attention projects $64{\rightarrow}256$ internally. Because the same $K{=}16$ tokens are appended for every slice, even the most distant slice attends to an identical whole-volume summary.
\begin{center}
\includegraphics[width=0.9\linewidth]{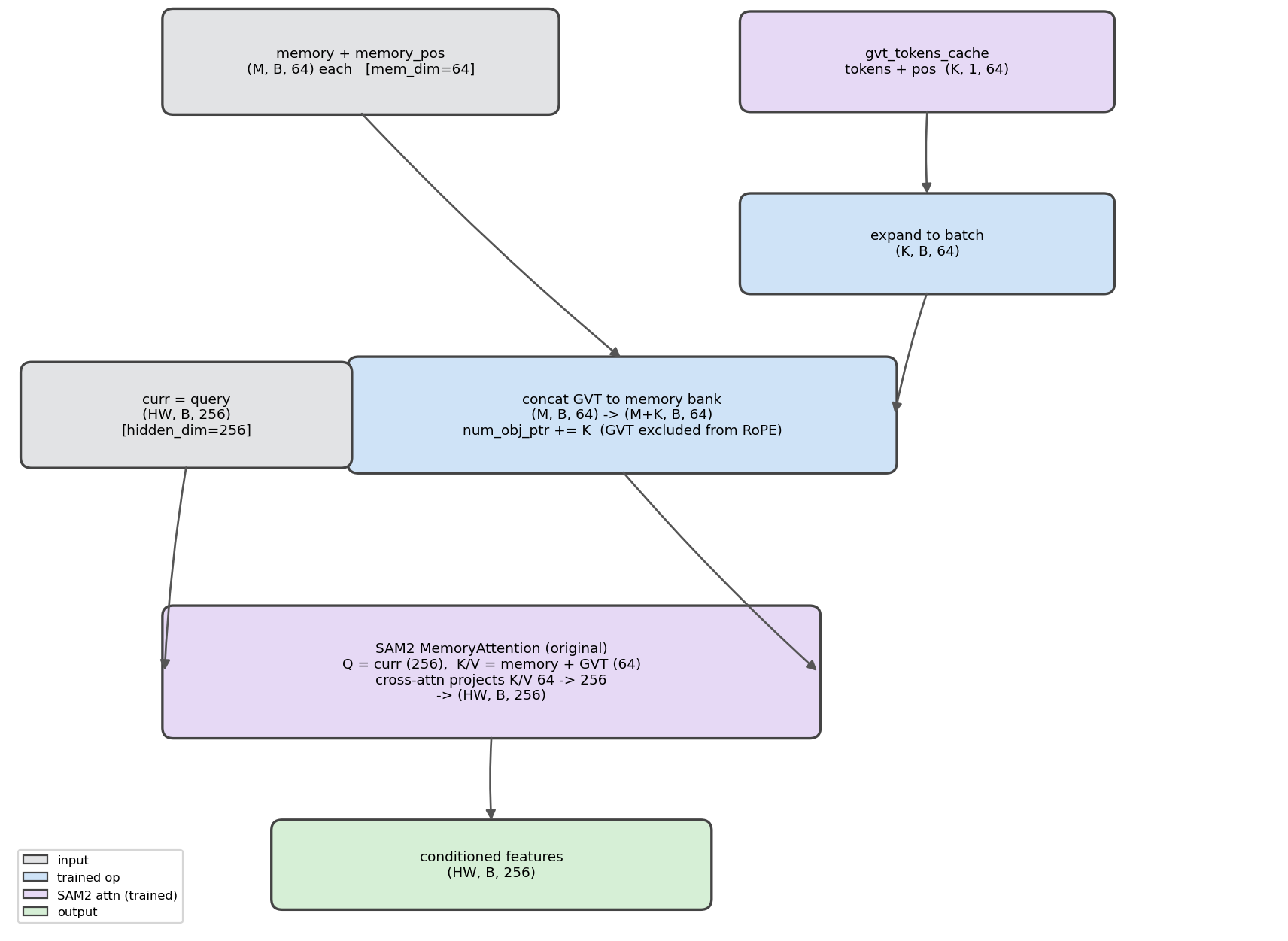}\\[3pt]
{\footnotesize\itshape Memory-bank fusion: the $K$ tokens are concatenated onto SAM2s memory bank before the original memory attention.}
\end{center}

\noindent\textbf{TSDF decoder (training only).} For each label and slice, the $K$ GVT tokens (each given an additive learned slice positional embedding) cross-attend to the prompted-slice memory features ($4096\times64$). The conditioned tokens are flattened ($16\times64{=}1024$) and linearly projected to a $64\times4\times4$ spatial seed, then upsampled by six ConvTranspose2d stages (channels $64{\to}64{\to}64{\to}32{\to}32{\to}16{\to}8$; spatial $4{\to}256$, each with GroupNorm and GELU) and a final $1{\times}1$ convolution to a single-channel $256\times256$ TSDF (linear output). It has ${\approx}30$K parameters and is discarded at inference.
\begin{center}
\includegraphics[width=0.6\linewidth]{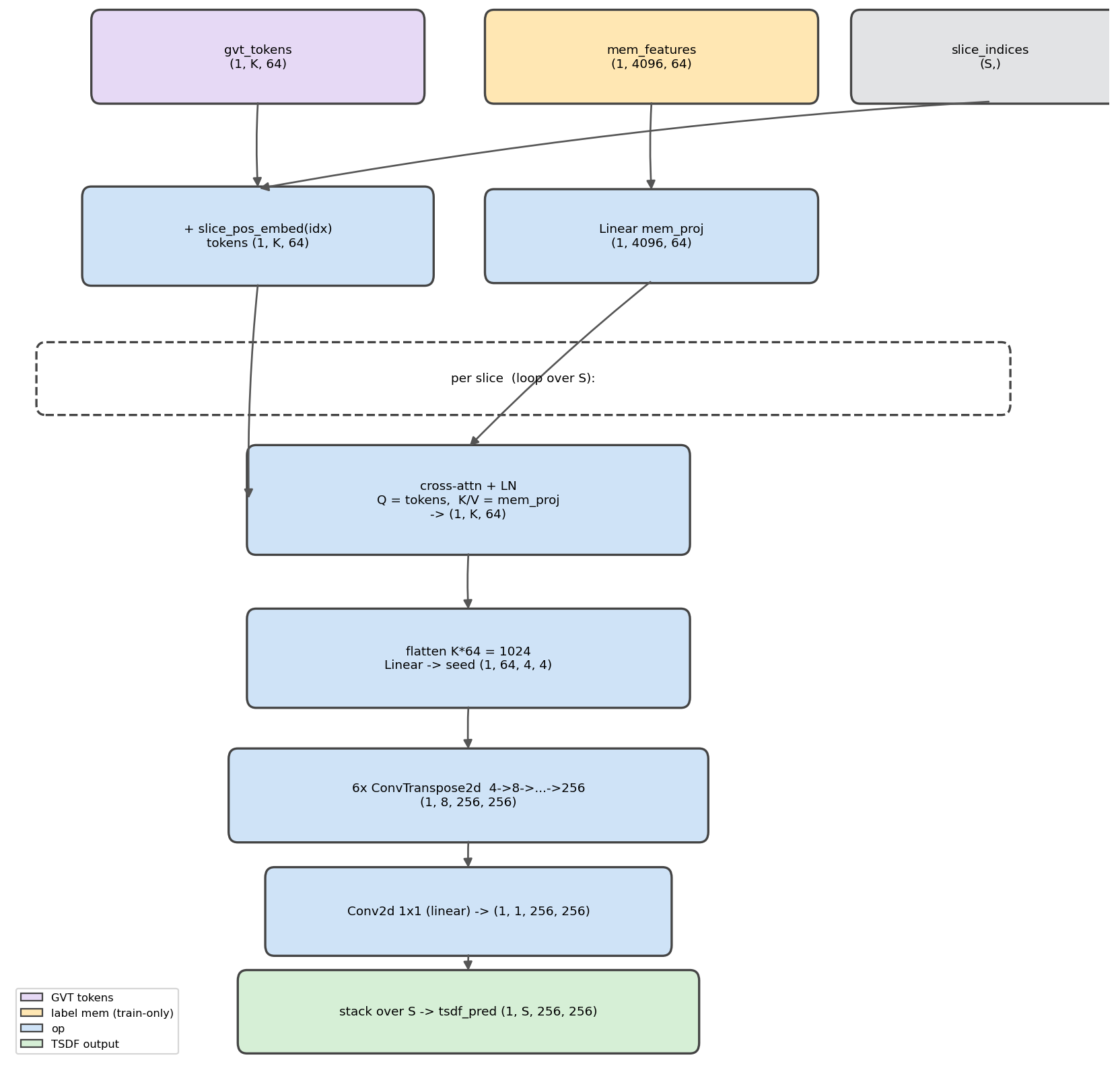}\\[3pt]
{\footnotesize\itshape TSDF decoder (training only): per-slice cross-attention conditioning then a six-stage $4{
ightarrow}256$ ConvTranspose stack.}
\end{center}

\noindent\textbf{Trainable and frozen components.} The image and prompt encoders are frozen; only the lightweight decoder, memory, and GVT modules are trained.

\begin{center}{\footnotesize\setlength{\tabcolsep}{4pt}
\begin{tabular}{|l|c|c|}
\hline
\textbf{Component} & \textbf{Status} & \textbf{Params} \\ \hline\hline
Hiera image encoder & Frozen & --- \\
Prompt encoder & Frozen & --- \\
Mask decoder & Trained & SAM2 \\
Memory encoder & Trained & SAM2 \\
Memory attention & Trained & SAM2 \\
GVT encoder & Trained & $\approx$ 120K \\
TSDF decoder & Trained; discarded at inference & $\approx$ 30K \\
\hline
\end{tabular}}\end{center}

\noindent\textbf{Training.} Models are trained from the pre-computed float16 embeddings with AdamW (learning rate $5\times10^{-6}$, weight decay 0.01), batch size 2, and AMP mixed precision, for 50 epochs on three GPUs. Each volume uses an 80\%/20\% mix of volumetric sequences and 2D slice batches, with video processing capped at $\mathrm{max\_slices}=32$ per sub-sequence. The TSDF objective uses $\tau=32$~px, $\alpha=4$, and $\sigma=\tau/3\approx10.7$.

\subsection*{C.\quad GVT Ablation Strategies}
\begin{center}
\includegraphics[width=0.95\linewidth]{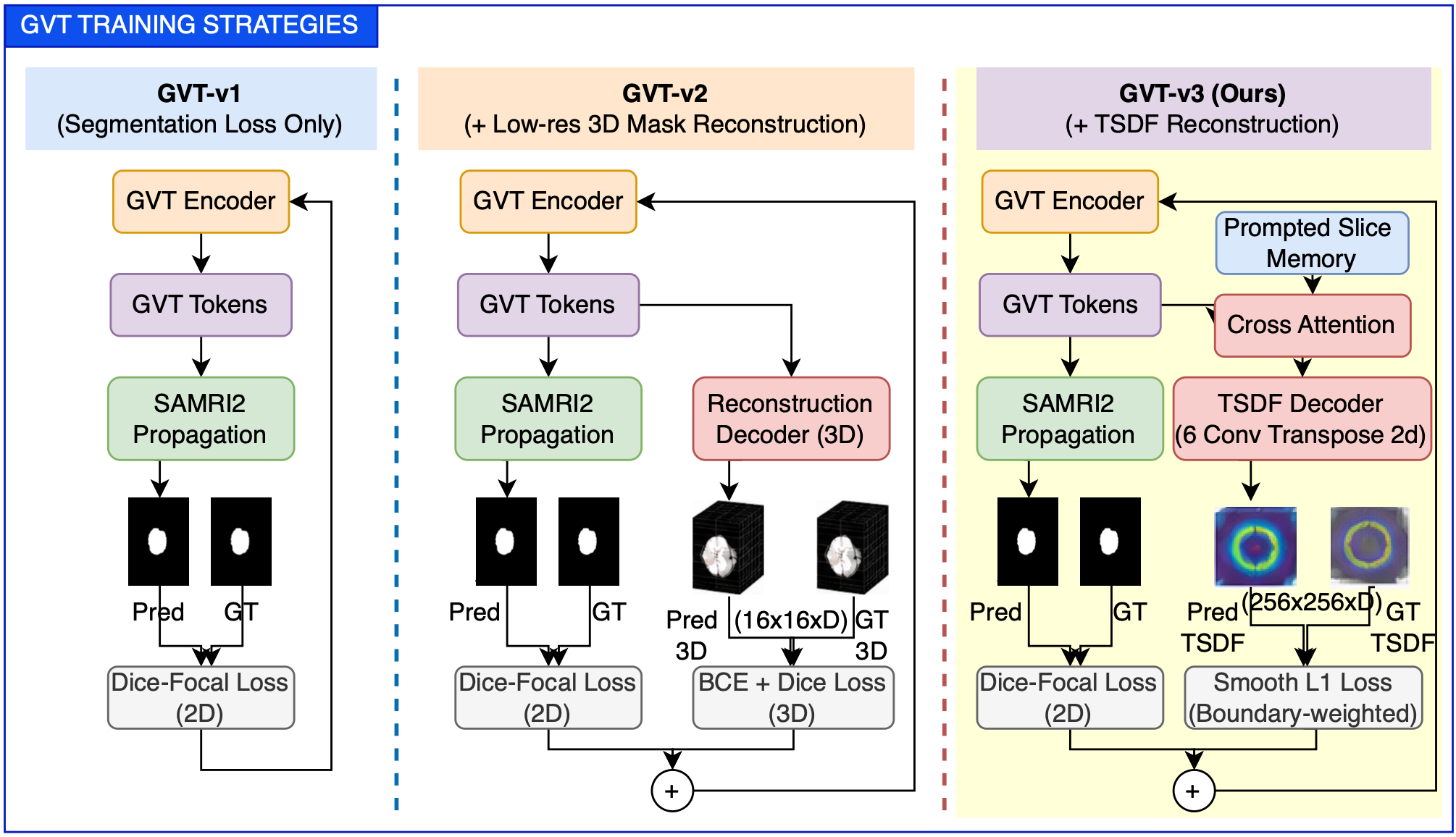}\\[3pt]
{\footnotesize\itshape \textbf{Figure S1.} GVT ablation strategies. GVT-v1 trains the tokens through the SAM2 segmentation loss only; GVT-v2 adds a low-resolution binary-mask reconstruction; GVT-v3 replaces it with a Truncated Signed Distance Field (TSDF) reconstruction. The auxiliary decoder is discarded at inference.}
\end{center}

\subsection*{D.\quad Model Scaling and Efficiency}
\medskip\noindent{\footnotesize\textbf{Table S1.} SAM2 zero-shot performance by model size.}\par\nobreak\smallskip

\begin{center}{\footnotesize\setlength{\tabcolsep}{4pt}
\begin{tabular}{|l|c|c|}
\hline
\textbf{Model} & \textbf{Params} & \textbf{Mean Dice} \\ \hline\hline
sam2\_tiny & 38.9M & 0.571 \\
sam2\_small & 46.0M & 0.583 \\
sam2\_b+ & 80.8M & 0.564 \\
sam2\_large & 224.4M & 0.558 \\
\hline
\end{tabular}}\end{center}

\medskip\noindent{\footnotesize\textbf{Table S2.} Efficiency comparison of GVT variants (parameter overhead).}\par\nobreak\smallskip

\begin{center}{\footnotesize\setlength{\tabcolsep}{4pt}
\begin{tabular}{|l|c|c|c|}
\hline
\textbf{Variant} & \textbf{GVT Params} & \textbf{Aux Decoder} & \textbf{Total Overhead} \\ \hline\hline
Baseline & -- & -- & 0 \\
GVT-v1 & ~120K & -- & <0.9\% \\
GVT-v2 & ~120K & ~85K & <1.5\% \\
SAMRI-3D & ~120K & ~30K & <1.1\% \\
\hline
\end{tabular}}\end{center}

\subsection*{E.\quad Per-Domain and Per-Sequence}
\medskip\noindent{\footnotesize\textbf{Table S3.} Per-domain Dice (mean $\pm$ SD of per-dataset median Dice across the 12 anatomical domains).}\par\nobreak\smallskip

\begin{center}{\footnotesize\setlength{\tabcolsep}{4pt}
\begin{tabular}{|l|c|c|c|c|}
\hline
\textbf{Domain} & \textbf{n} & \textbf{SAM2-Small} & \textbf{SAMed-2} & \textbf{SAMRI-3D} \\ \hline\hline
Brain & 8 & 0.44 $\pm$ 0.29 & 0.60 $\pm$ 0.16 & 0.67 $\pm$ 0.17 \\
Prostate & 7 & 0.74 $\pm$ 0.15 & 0.84 $\pm$ 0.10 & 0.85 $\pm$ 0.11 \\
Knee/MSK & 8 & 0.65 $\pm$ 0.23 & 0.68 $\pm$ 0.05 & 0.83 $\pm$ 0.09 \\
Cardiac & 3 & 0.68 $\pm$ 0.21 & 0.80 $\pm$ 0.02 & 0.87 $\pm$ 0.02 \\
Whole-body & 1 & 0.23 $\pm$ 0.15 & 0.66 $\pm$ 0.16 & 0.57 $\pm$ 0.13 \\
Hippocampus & 1 & 0.65 $\pm$ 0.13 & 0.74 $\pm$ 0.04 & 0.84 $\pm$ 0.05 \\
Vestibular & 1 & 0.34 $\pm$ 0.11 & 0.45 $\pm$ 0.05 & 0.63 $\pm$ 0.09 \\
Hip & 1 & 0.55 $\pm$ 0.21 & 0.74 $\pm$ 0.03 & 0.87 $\pm$ 0.04 \\
Spine & 1 & 0.78 $\pm$ 0.07 & 0.65 $\pm$ 0.02 & 0.84 $\pm$ 0.02 \\
Abdominal & 1 & 0.28 $\pm$ 0.10 & 0.72 $\pm$ 0.04 & 0.70 $\pm$ 0.10 \\
Shoulder & 1 & 0.36 $\pm$ 0.16 & 0.27 $\pm$ 0.06 & 0.72 $\pm$ 0.18 \\
Gynecology & 1 & 0.67 $\pm$ 0.00 & 0.68 $\pm$ 0.00 & 0.76 $\pm$ 0.00 \\
\hline
\end{tabular}}\end{center}

\medskip\noindent{\footnotesize\textbf{Table S4.} Per-sequence Dice (mean $\pm$ SD of per-dataset median Dice across datasets with each sequence).}\par\nobreak\smallskip

\begin{center}{\footnotesize\setlength{\tabcolsep}{4pt}
\begin{tabular}{|l|c|c|c|c|c|}
\hline
\textbf{Sequence} & \textbf{n} & \textbf{SAM2-Small} & \textbf{SAMed-2} & \textbf{SAMRI-3D-2dloss} & \textbf{SAMRI-3D} \\ \hline\hline
T1 & 4 & 0.43 $\pm$ 0.30 & 0.65 $\pm$ 0.18 & 0.71 $\pm$ 0.16 & 0.72 $\pm$ 0.13 \\
T1CE & 2 & 0.51 $\pm$ 0.23 & 0.46 $\pm$ 0.01 & 0.71 $\pm$ 0.16 & 0.72 $\pm$ 0.12 \\
FLAIR & 3 & 0.39 $\pm$ 0.37 & 0.62 $\pm$ 0.20 & 0.60 $\pm$ 0.28 & 0.61 $\pm$ 0.25 \\
DWI & 1 & 0.58 $\pm$ 0.22 & 0.73 $\pm$ 0.06 & 0.81 $\pm$ 0.15 & 0.80 $\pm$ 0.13 \\
ADC & 3 & 0.56 $\pm$ 0.26 & 0.77 $\pm$ 0.15 & 0.72 $\pm$ 0.21 & 0.74 $\pm$ 0.17 \\
T2 & 10 & 0.70 $\pm$ 0.17 & 0.72 $\pm$ 0.19 & 0.82 $\pm$ 0.12 & 0.84 $\pm$ 0.09 \\
Cine & 1 & 0.85 $\pm$ 0.07 & 0.79 $\pm$ 0.06 & 0.85 $\pm$ 0.07 & 0.85 $\pm$ 0.06 \\
DESS & 3 & 0.75 $\pm$ 0.04 & 0.67 $\pm$ 0.05 & 0.90 $\pm$ 0.02 & 0.89 $\pm$ 0.03 \\
TSE & 1 & 0.55 $\pm$ 0.08 & 0.59 $\pm$ 0.03 & 0.72 $\pm$ 0.08 & 0.70 $\pm$ 0.08 \\
FLASH & 2 & 0.38 $\pm$ 0.37 & 0.71 $\pm$ 0.04 & 0.58 $\pm$ 0.28 & 0.75 $\pm$ 0.06 \\
PD & 1 & 0.79 $\pm$ 0.11 & 0.71 $\pm$ 0.00 & 0.90 $\pm$ 0.01 & 0.88 $\pm$ 0.02 \\
\hline
\end{tabular}}\end{center}

\subsection*{F.\quad Benchmark Datasets}
\medskip\noindent{\footnotesize\textbf{Table S5.} Complete list of the 34 SAMRI-3D benchmark datasets.}\par\nobreak\smallskip

{\scriptsize\setlength{\tabcolsep}{4pt}\setlength\LTleft{0pt}\setlength\LTright{0pt}
\begin{longtable}{|@{\extracolsep{\fill}}l|l|c|c|c|c|}
\hline
\textbf{Domain} & \textbf{Dataset} & \textbf{Sequence} & \textbf{\# Vol.} & \textbf{\# Labels} & \textbf{Split} \\ \hline\endfirsthead
\hline \textbf{Domain} & \textbf{Dataset} & \textbf{Sequence} & \textbf{\# Vol.} & \textbf{\# Labels} & \textbf{Split} \\ \hline\endhead
Brain & MSD-BrainTumour [1] & FLAIR & 1,936 & 3 & Seen \\
Brain & BraTS-FLAIR [2] & FLAIR & 954 & 1 & Seen \\
Brain & BraTS-T1 [2] & T1w & 954 & 1 & Seen \\
Brain & BraTS-T1CE [2] & T1CE & 890 & 1 & Seen \\
Brain & ISLES2022-DWI [3] & DWI & 224 & 1 & Seen \\
Brain & ISLES2022-ADC [3] & ADC & 224 & 1 & Seen \\
Brain & WMH-T1 [4] & T1w & 169 & 3 & Seen \\
Brain & WMH-FLAIR [4] & FLAIR & 169 & 3 & Seen \\
\hline
\end{longtable}}

\subsection*{G.\quad Per-Dataset and Per-Target Performance}
\begin{landscape}
\medskip\noindent{\footnotesize\textbf{Table S6.} DSC by dataset --- median (Q1, Q3) across volumes; final row is mean $\pm$ SD.}\par\nobreak\smallskip

{\tiny\setlength{\tabcolsep}{3pt}\setlength\LTleft{-0.75in}\setlength\LTright{-0.75in}
\begin{longtable}{|@{\extracolsep{\fill}}l|l|c|c|c|c|c|c|c|c|c|c|c|}
\hline
\textbf{Domain} & \textbf{Dataset} & \textbf{SAM2-Tiny} & \textbf{SAM2-Small} & \textbf{SAM2-Base+} & \textbf{SAM2-Large} & \textbf{Medical-SAM2} & \textbf{SAM-Med3D} & \textbf{SAMed-2} & \textbf{SAMRI-3D-2dloss} & \textbf{GVT-v1} & \textbf{GVT-v2} & \textbf{SAMRI-3D} \\ \hline\endfirsthead
\hline \textbf{Domain} & \textbf{Dataset} & \textbf{SAM2-Tiny} & \textbf{SAM2-Small} & \textbf{SAM2-Base+} & \textbf{SAM2-Large} & \textbf{Medical-SAM2} & \textbf{SAM-Med3D} & \textbf{SAMed-2} & \textbf{SAMRI-3D-2dloss} & \textbf{GVT-v1} & \textbf{GVT-v2} & \textbf{SAMRI-3D} \\ \hline\endhead
Brain & MSD-BrainTumour & 0.24 (0.05, 0.42) & 0.26 (0.05, 0.45) & 0.23 (0.05, 0.44) & 0.21 (0.04, 0.40) & 0.23 (0.10, 0.38) & 0.34 (0.17, 0.53) & 0.57 (0.39, 0.65) & 0.33 (0.07, 0.53) & 0.40 (0.11, 0.60) & 0.31 (0.08, 0.54) & 0.40 (0.10, 0.61) \\
Brain & BraTS-FLAIR & 0.82 (0.73, 0.88) & 0.81 (0.72, 0.87) & 0.84 (0.75, 0.88) & 0.83 (0.72, 0.88) & 0.71 (0.60, 0.78) & 0.76 (0.70, 0.83) & 0.84 (0.79, 0.87) & 0.89 (0.82, 0.92) & 0.89 (0.83, 0.91) & 0.90 (0.83, 0.92) & 0.89 (0.82, 0.92) \\
Brain & BraTS-T1 & 0.71 (0.64, 0.77) & 0.72 (0.63, 0.76) & 0.71 (0.61, 0.77) & 0.68 (0.59, 0.75) & 0.59 (0.48, 0.67) & 0.60 (0.46, 0.70) & 0.76 (0.71, 0.80) & 0.79 (0.73, 0.84) & 0.79 (0.73, 0.84) & 0.81 (0.75, 0.86) & 0.80 (0.75, 0.84) \\
Brain & BraTS-T1CE & 0.66 (0.53, 0.80) & 0.67 (0.50, 0.80) & 0.66 (0.54, 0.79) & 0.67 (0.52, 0.80) & 0.60 (0.49, 0.76) & 0.28 (0.20, 0.40) & 0.47 (0.34, 0.62) & 0.82 (0.60, 0.90) & 0.86 (0.66, 0.91) & 0.86 (0.66, 0.91) & 0.80 (0.55, 0.89) \\
Brain & ISLES2022-DWI & 0.56 (0.33, 0.83) & 0.58 (0.28, 0.75) & 0.63 (0.47, 0.78) & 0.54 (0.22, 0.78) & 0.60 (0.43, 0.73) & 0.06 (0.03, 0.18) & 0.73 (0.65, 0.78) & 0.81 (0.61, 0.86) & 0.77 (0.60, 0.86) & 0.80 (0.64, 0.88) & 0.80 (0.70, 0.88) \\
Brain & ISLES2022-ADC & 0.40 (0.14, 0.59) & 0.33 (0.13, 0.60) & 0.43 (0.18, 0.59) & 0.33 (0.12, 0.61) & 0.37 (0.30, 0.52) & 0.07 (0.03, 0.15) & 0.60 (0.56, 0.67) & 0.51 (0.42, 0.71) & 0.58 (0.38, 0.72) & 0.55 (0.42, 0.73) & 0.59 (0.45, 0.71) \\
Brain & WMH-T1 & 0.07 (0.03, 0.14) & 0.08 (0.04, 0.17) & 0.13 (0.06, 0.29) & 0.09 (0.04, 0.17) & 0.38 (0.25, 0.48) & 0.02 (0.01, 0.04) & 0.38 (0.29, 0.50) & 0.48 (0.32, 0.61) & 0.43 (0.25, 0.61) & 0.53 (0.40, 0.64) & 0.54 (0.40, 0.65) \\
Brain & WMH-FLAIR & 0.13 (0.04, 0.29) & 0.11 (0.04, 0.24) & 0.17 (0.08, 0.44) & 0.15 (0.05, 0.31) & 0.42 (0.25, 0.53) & 0.02 (0.01, 0.06) & 0.45 (0.36, 0.55) & 0.59 (0.39, 0.71) & 0.46 (0.25, 0.66) & 0.58 (0.38, 0.73) & 0.54 (0.41, 0.72) \\
\hline
\end{longtable}}
\end{landscape}

\begin{landscape}
\medskip\noindent{\footnotesize\textbf{Table S7.} HD by dataset --- median (Q1, Q3) across volumes; final row is mean $\pm$ SD.}\par\nobreak\smallskip

{\tiny\setlength{\tabcolsep}{3pt}\setlength\LTleft{-0.75in}\setlength\LTright{-0.75in}
\begin{longtable}{|@{\extracolsep{\fill}}l|l|c|c|c|c|c|c|c|c|c|c|c|}
\hline
\textbf{Domain} & \textbf{Dataset} & \textbf{SAM2-Tiny} & \textbf{SAM2-Small} & \textbf{SAM2-Base+} & \textbf{SAM2-Large} & \textbf{Medical-SAM2} & \textbf{SAM-Med3D} & \textbf{SAMed-2} & \textbf{SAMRI-3D-2dloss} & \textbf{GVT-v1} & \textbf{GVT-v2} & \textbf{SAMRI-3D} \\ \hline\endfirsthead
\hline \textbf{Domain} & \textbf{Dataset} & \textbf{SAM2-Tiny} & \textbf{SAM2-Small} & \textbf{SAM2-Base+} & \textbf{SAM2-Large} & \textbf{Medical-SAM2} & \textbf{SAM-Med3D} & \textbf{SAMed-2} & \textbf{SAMRI-3D-2dloss} & \textbf{GVT-v1} & \textbf{GVT-v2} & \textbf{SAMRI-3D} \\ \hline\endhead
Brain & MSD-BrainTumour & 63.90 (53.55, 76.05) & 59.77 (43.15, 72.10) & 57.22 (41.21, 71.12) & 65.72 (50.28, 77.99) & 48.43 (32.16, 62.45) & 21.47 (15.02, 29.70) & 14.46 (9.26, 21.16) & 35.28 (21.22, 49.45) & 26.48 (17.81, 41.86) & 40.56 (23.40, 64.26) & 38.41 (21.62, 52.73) \\
Brain & BraTS-FLAIR & 22.19 (14.04, 30.95) & 24.51 (15.69, 37.59) & 17.84 (11.79, 34.77) & 24.93 (16.06, 40.72) & 30.13 (20.17, 39.99) & 13.96 (10.92, 19.40) & 19.44 (11.57, 53.37) & 14.05 (8.77, 24.62) & 14.92 (9.68, 22.60) & 13.25 (8.46, 25.81) & 13.38 (8.53, 21.85) \\
Brain & BraTS-T1 & 24.71 (17.51, 35.07) & 29.91 (19.80, 37.10) & 29.05 (20.47, 38.46) & 34.43 (26.21, 51.98) & 27.08 (23.33, 34.02) & 20.00 (15.57, 26.21) & 17.01 (11.97, 21.45) & 18.07 (13.42, 26.81) & 18.03 (14.40, 27.24) & 19.40 (12.61, 31.76) & 16.76 (12.78, 23.19) \\
Brain & BraTS-T1CE & 8.80 (6.32, 11.58) & 9.27 (6.65, 13.38) & 8.80 (6.71, 12.44) & 8.63 (6.10, 12.28) & 9.77 (7.35, 16.26) & 11.18 (9.43, 15.20) & 8.15 (6.16, 12.53) & 7.62 (5.41, 10.68) & 7.07 (5.20, 10.63) & 7.45 (5.02, 10.04) & 7.75 (5.54, 11.52) \\
Brain & ISLES2022-DWI & 15.78 (8.31, 26.40) & 17.42 (10.46, 25.51) & 11.49 (4.36, 24.10) & 19.10 (9.50, 26.94) & 5.61 (2.24, 9.84) & 16.03 (13.03, 19.12) & 2.83 (2.00, 4.21) & 3.74 (1.49, 7.45) & 3.87 (2.06, 7.88) & 3.24 (1.41, 5.17) & 3.32 (1.73, 5.92) \\
Brain & ISLES2022-ADC & 16.58 (5.10, 26.60) & 20.55 (8.00, 24.57) & 12.08 (4.86, 18.69) & 22.13 (8.97, 29.00) & 10.03 (5.49, 17.22) & 16.36 (14.01, 19.67) & 3.16 (2.40, 5.02) & 4.69 (2.93, 8.65) & 4.58 (3.12, 8.80) & 4.47 (2.96, 7.98) & 4.24 (2.45, 7.87) \\
Brain & WMH-T1 & 45.84 (32.75, 62.95) & 44.37 (32.83, 66.59) & 32.77 (14.46, 53.24) & 50.98 (33.83, 68.88) & 6.70 (3.32, 16.44) & 16.42 (14.06, 20.55) & 5.00 (3.61, 8.99) & 6.78 (3.46, 11.26) & 8.60 (3.94, 14.11) & 4.53 (2.40, 8.08) & 4.58 (2.45, 9.26) \\
\hline
\end{longtable}}
\end{landscape}

\begin{landscape}
\medskip\noindent{\footnotesize\textbf{Table S8.} MSD by dataset --- median (Q1, Q3) across volumes; final row is mean $\pm$ SD.}\par\nobreak\smallskip

{\tiny\setlength{\tabcolsep}{3pt}\setlength\LTleft{-0.75in}\setlength\LTright{-0.75in}
\begin{longtable}{|@{\extracolsep{\fill}}l|l|c|c|c|c|c|c|c|c|c|c|c|}
\hline
\textbf{Domain} & \textbf{Dataset} & \textbf{SAM2-Tiny} & \textbf{SAM2-Small} & \textbf{SAM2-Base+} & \textbf{SAM2-Large} & \textbf{Medical-SAM2} & \textbf{SAM-Med3D} & \textbf{SAMed-2} & \textbf{SAMRI-3D-2dloss} & \textbf{GVT-v1} & \textbf{GVT-v2} & \textbf{SAMRI-3D} \\ \hline\endfirsthead
\hline \textbf{Domain} & \textbf{Dataset} & \textbf{SAM2-Tiny} & \textbf{SAM2-Small} & \textbf{SAM2-Base+} & \textbf{SAM2-Large} & \textbf{Medical-SAM2} & \textbf{SAM-Med3D} & \textbf{SAMed-2} & \textbf{SAMRI-3D-2dloss} & \textbf{GVT-v1} & \textbf{GVT-v2} & \textbf{SAMRI-3D} \\ \hline\endhead
Brain & MSD-BrainTumour & 12.26 (8.70, 16.41) & 10.54 (6.99, 14.59) & 9.89 (6.74, 14.34) & 11.50 (7.99, 16.33) & 8.85 (6.02, 12.52) & 4.08 (3.02, 5.87) & 2.48 (2.04, 3.21) & 6.39 (4.00, 9.04) & 4.92 (3.22, 7.06) & 6.30 (3.94, 9.23) & 5.67 (3.55, 8.27) \\
Brain & BraTS-FLAIR & 3.32 (2.54, 4.55) & 3.50 (2.58, 4.97) & 3.18 (2.23, 4.36) & 3.26 (2.20, 4.94) & 4.67 (3.48, 5.96) & 2.36 (2.05, 2.79) & 2.42 (2.08, 2.91) & 2.24 (1.74, 2.94) & 2.24 (1.70, 3.02) & 1.93 (1.60, 2.64) & 2.07 (1.65, 3.07) \\
Brain & BraTS-T1 & 4.45 (3.52, 5.82) & 4.64 (3.73, 6.01) & 4.83 (3.60, 6.04) & 5.23 (3.81, 7.34) & 5.50 (4.44, 7.11) & 4.13 (3.57, 5.23) & 3.02 (2.60, 3.45) & 3.29 (2.60, 4.29) & 3.46 (2.58, 4.32) & 2.88 (2.46, 3.75) & 3.31 (2.49, 4.02) \\
Brain & BraTS-T1CE & 2.32 (1.67, 3.18) & 2.62 (1.68, 3.27) & 2.31 (1.69, 3.31) & 2.37 (1.78, 3.29) & 2.48 (1.94, 3.13) & 3.14 (2.70, 3.73) & 1.90 (1.71, 2.22) & 1.68 (1.32, 2.46) & 1.55 (1.26, 2.15) & 1.47 (1.25, 2.40) & 1.67 (1.32, 2.74) \\
Brain & ISLES2022-DWI & 3.72 (1.61, 6.17) & 3.56 (2.20, 6.30) & 3.12 (1.37, 5.87) & 4.95 (1.86, 6.41) & 1.51 (1.13, 2.98) & 6.25 (4.78, 7.63) & 1.16 (1.10, 1.26) & 1.29 (1.04, 1.82) & 1.32 (1.07, 1.82) & 1.17 (1.03, 1.51) & 1.27 (1.07, 1.59) \\
Brain & ISLES2022-ADC & 3.25 (1.49, 5.49) & 4.26 (1.97, 6.50) & 2.88 (1.57, 4.23) & 3.85 (2.05, 6.62) & 2.82 (1.80, 3.64) & 5.28 (4.13, 6.66) & 1.22 (1.16, 1.40) & 1.49 (1.17, 2.33) & 1.60 (1.18, 2.22) & 1.49 (1.19, 2.45) & 1.41 (1.17, 2.21) \\
Brain & WMH-T1 & 9.01 (5.90, 12.37) & 8.30 (5.70, 12.39) & 6.72 (3.45, 11.17) & 9.91 (6.22, 15.05) & 2.19 (1.32, 4.84) & 6.90 (5.94, 7.81) & 1.55 (1.33, 2.09) & 1.85 (1.30, 2.62) & 2.04 (1.34, 3.26) & 1.48 (1.18, 2.08) & 1.49 (1.22, 2.30) \\
Brain & WMH-FLAIR & 9.79 (5.72, 17.75) & 8.85 (5.37, 13.06) & 7.09 (3.41, 14.01) & 10.20 (6.08, 16.12) & 2.21 (1.30, 4.55) & 6.84 (5.70, 7.94) & 1.67 (1.35, 2.08) & 1.67 (1.23, 3.39) & 2.77 (1.43, 5.11) & 1.63 (1.18, 2.92) & 1.57 (1.19, 2.79) \\
\hline
\end{longtable}}
\end{landscape}

\begin{landscape}
\medskip\noindent{\footnotesize\textbf{Table S9.} DSC by target --- median (Q1, Q3) across volumes; final row is mean $\pm$ SD.}\par\nobreak\smallskip

{\tiny\setlength{\tabcolsep}{3pt}\setlength\LTleft{-0.75in}\setlength\LTright{-0.75in}
\begin{longtable}{|@{\extracolsep{\fill}}l|l|c|c|c|c|c|c|c|c|c|c|c|}
\hline
\textbf{Dataset} & \textbf{Target} & \textbf{SAM2-Tiny} & \textbf{SAM2-Small} & \textbf{SAM2-Base+} & \textbf{SAM2-Large} & \textbf{Medical-SAM2} & \textbf{SAM-Med3D} & \textbf{SAMed-2} & \textbf{SAMRI-3D-2dloss} & \textbf{GVT-v1} & \textbf{GVT-v2} & \textbf{SAMRI-3D} \\ \hline\endfirsthead
\hline \textbf{Dataset} & \textbf{Target} & \textbf{SAM2-Tiny} & \textbf{SAM2-Small} & \textbf{SAM2-Base+} & \textbf{SAM2-Large} & \textbf{Medical-SAM2} & \textbf{SAM-Med3D} & \textbf{SAMed-2} & \textbf{SAMRI-3D-2dloss} & \textbf{GVT-v1} & \textbf{GVT-v2} & \textbf{SAMRI-3D} \\ \hline\endhead
MSD-BrainTumour & Edema & 0.31 (0.17,0.49) & 0.33 (0.19,0.49) & 0.30 (0.15,0.46) & 0.29 (0.16,0.45) & 0.24 (0.12,0.38) & 0.43 (0.27,0.62) & 0.57 (0.44,0.63) & 0.35 (0.19,0.53) & 0.44 (0.27,0.61) & 0.39 (0.22,0.57) & 0.45 (0.28,0.62) \\
MSD-BrainTumour & Enhancing tumour & 0.35 (0.23,0.52) & 0.40 (0.27,0.55) & 0.42 (0.24,0.59) & 0.35 (0.22,0.51) & 0.25 (0.11,0.39) & 0.39 (0.17,0.55) & 0.63 (0.55,0.69) & 0.51 (0.39,0.63) & 0.57 (0.44,0.68) & 0.52 (0.36,0.62) & 0.56 (0.43,0.69) \\
MSD-BrainTumour & Non-enhancing tumor & 0.02 (0.00,0.08) & 0.03 (0.00,0.08) & 0.02 (0.00,0.09) & 0.02 (0.00,0.06) & 0.19 (0.08,0.36) & 0.22 (0.12,0.37) & 0.44 (0.30,0.61) & 0.03 (0.00,0.11) & 0.05 (0.00,0.12) & 0.04 (0.01,0.11) & 0.06 (0.01,0.18) \\
BraTS-FLAIR & Brain Tumor & 0.82 (0.73,0.88) & 0.81 (0.72,0.87) & 0.84 (0.75,0.88) & 0.83 (0.72,0.88) & 0.71 (0.60,0.78) & 0.76 (0.70,0.83) & 0.84 (0.79,0.87) & 0.89 (0.82,0.92) & 0.89 (0.83,0.91) & 0.90 (0.83,0.92) & 0.89 (0.82,0.92) \\
BraTS-T1 & Brain Tumor & 0.71 (0.64,0.77) & 0.72 (0.63,0.76) & 0.71 (0.61,0.77) & 0.68 (0.59,0.75) & 0.59 (0.48,0.67) & 0.60 (0.46,0.70) & 0.76 (0.71,0.80) & 0.79 (0.73,0.84) & 0.79 (0.73,0.84) & 0.81 (0.75,0.86) & 0.80 (0.75,0.84) \\
BraTS-T1CE & Brain Tumor & 0.66 (0.53,0.80) & 0.67 (0.50,0.80) & 0.66 (0.54,0.79) & 0.67 (0.52,0.80) & 0.60 (0.49,0.76) & 0.28 (0.20,0.40) & 0.47 (0.34,0.62) & 0.82 (0.60,0.90) & 0.86 (0.66,0.91) & 0.86 (0.66,0.91) & 0.80 (0.55,0.89) \\
ISLES2022-DWI & Ischemic Stroke & 0.56 (0.33,0.83) & 0.58 (0.28,0.75) & 0.63 (0.47,0.78) & 0.54 (0.22,0.78) & 0.60 (0.43,0.73) & 0.06 (0.03,0.18) & 0.73 (0.65,0.78) & 0.81 (0.61,0.86) & 0.77 (0.60,0.86) & 0.80 (0.64,0.88) & 0.80 (0.70,0.88) \\
ISLES2022-ADC & Ischemic Stroke & 0.40 (0.14,0.59) & 0.33 (0.13,0.60) & 0.43 (0.18,0.59) & 0.33 (0.12,0.61) & 0.37 (0.30,0.52) & 0.07 (0.03,0.15) & 0.60 (0.56,0.67) & 0.51 (0.42,0.71) & 0.58 (0.38,0.72) & 0.55 (0.42,0.73) & 0.59 (0.45,0.71) \\
\hline
\end{longtable}}
\end{landscape}

\begin{landscape}
\medskip\noindent{\footnotesize\textbf{Table S10.} HD by target --- median (Q1, Q3) across volumes; final row is mean $\pm$ SD.}\par\nobreak\smallskip

{\tiny\setlength{\tabcolsep}{3pt}\setlength\LTleft{-0.75in}\setlength\LTright{-0.75in}
\begin{longtable}{|@{\extracolsep{\fill}}l|l|c|c|c|c|c|c|c|c|c|c|c|}
\hline
\textbf{Dataset} & \textbf{Target} & \textbf{SAM2-Tiny} & \textbf{SAM2-Small} & \textbf{SAM2-Base+} & \textbf{SAM2-Large} & \textbf{Medical-SAM2} & \textbf{SAM-Med3D} & \textbf{SAMed-2} & \textbf{SAMRI-3D-2dloss} & \textbf{GVT-v1} & \textbf{GVT-v2} & \textbf{SAMRI-3D} \\ \hline\endfirsthead
\hline \textbf{Dataset} & \textbf{Target} & \textbf{SAM2-Tiny} & \textbf{SAM2-Small} & \textbf{SAM2-Base+} & \textbf{SAM2-Large} & \textbf{Medical-SAM2} & \textbf{SAM-Med3D} & \textbf{SAMed-2} & \textbf{SAMRI-3D-2dloss} & \textbf{GVT-v1} & \textbf{GVT-v2} & \textbf{SAMRI-3D} \\ \hline\endhead
MSD-BrainTumour & Edema & 58.89 (49.18,68.89) & 54.80 (43.02,66.20) & 54.90 (43.38,65.55) & 61.06 (48.31,72.89) & 48.49 (36.74,57.79) & 23.26 (15.66,31.08) & 19.25 (14.27,24.19) & 42.28 (29.72,52.45) & 32.65 (21.37,46.05) & 49.13 (31.14,74.48) & 42.30 (27.20,52.34) \\
MSD-BrainTumour & Enhancing tumour & 68.17 (57.78,78.93) & 63.11 (45.55,75.98) & 61.26 (37.55,73.72) & 70.81 (56.02,80.91) & 45.77 (26.41,65.17) & 20.86 (13.83,28.28) & 10.39 (7.86,17.27) & 27.25 (18.05,45.15) & 19.30 (11.67,30.59) & 31.96 (18.99,58.14) & 27.25 (16.05,53.46) \\
MSD-BrainTumour & Non-enhancing tumor & 65.68 (55.75,76.42) & 61.12 (41.43,72.63) & 57.12 (34.41,74.30) & 65.06 (45.42,78.22) & 49.49 (32.12,65.43) & 21.15 (13.99,29.38) & 11.87 (8.06,20.62) & 34.43 (22.05,49.00) & 30.81 (21.25,45.07) & 39.50 (26.01,64.78) & 38.48 (23.63,53.80) \\
BraTS-FLAIR & Brain Tumor & 22.19 (14.04,30.95) & 24.51 (15.69,37.59) & 17.84 (11.79,34.77) & 24.93 (16.06,40.72) & 30.22 (20.20,40.72) & 13.96 (10.92,19.40) & 19.44 (11.57,53.37) & 14.05 (8.77,24.62) & 14.92 (9.68,22.60) & 13.25 (8.46,25.81) & 13.38 (8.53,21.85) \\
BraTS-T1 & Brain Tumor & 24.71 (17.51,35.07) & 29.91 (19.80,37.10) & 29.05 (20.47,38.46) & 34.43 (26.21,51.98) & 27.08 (23.33,34.02) & 20.00 (15.57,26.21) & 17.01 (11.97,21.45) & 18.07 (13.42,26.81) & 18.03 (14.40,27.24) & 19.40 (12.61,31.76) & 16.76 (12.78,23.19) \\
BraTS-T1CE & Brain Tumor & 8.80 (6.32,11.58) & 9.27 (6.65,13.38) & 8.80 (6.71,12.44) & 8.63 (6.10,12.28) & 9.77 (7.35,16.26) & 11.18 (9.43,15.20) & 8.15 (6.16,12.53) & 7.62 (5.41,10.68) & 7.07 (5.20,10.63) & 7.45 (5.02,10.04) & 7.75 (5.54,11.52) \\
ISLES2022-DWI & Ischemic Stroke & 16.54 (8.47,26.74) & 17.42 (10.46,25.51) & 11.77 (4.54,26.05) & 19.10 (9.50,26.94) & 5.61 (2.24,9.84) & 16.03 (13.03,19.12) & 2.83 (2.00,4.21) & 3.74 (1.49,7.45) & 3.87 (2.06,7.88) & 3.24 (1.41,5.17) & 3.46 (1.80,5.98) \\
\hline
\end{longtable}}
\end{landscape}

\begin{landscape}
\medskip\noindent{\footnotesize\textbf{Table S11.} MSD by target --- median (Q1, Q3) across volumes; final row is mean $\pm$ SD.}\par\nobreak\smallskip

{\tiny\setlength{\tabcolsep}{3pt}\setlength\LTleft{-0.75in}\setlength\LTright{-0.75in}
\begin{longtable}{|@{\extracolsep{\fill}}l|l|c|c|c|c|c|c|c|c|c|c|c|}
\hline
\textbf{Dataset} & \textbf{Target} & \textbf{SAM2-Tiny} & \textbf{SAM2-Small} & \textbf{SAM2-Base+} & \textbf{SAM2-Large} & \textbf{Medical-SAM2} & \textbf{SAM-Med3D} & \textbf{SAMed-2} & \textbf{SAMRI-3D-2dloss} & \textbf{GVT-v1} & \textbf{GVT-v2} & \textbf{SAMRI-3D} \\ \hline\endfirsthead
\hline \textbf{Dataset} & \textbf{Target} & \textbf{SAM2-Tiny} & \textbf{SAM2-Small} & \textbf{SAM2-Base+} & \textbf{SAM2-Large} & \textbf{Medical-SAM2} & \textbf{SAM-Med3D} & \textbf{SAMed-2} & \textbf{SAMRI-3D-2dloss} & \textbf{GVT-v1} & \textbf{GVT-v2} & \textbf{SAMRI-3D} \\ \hline\endhead
MSD-BrainTumour & Edema & 9.46 (7.59,12.05) & 8.77 (6.69,11.26) & 8.76 (6.51,11.72) & 9.74 (6.94,12.21) & 9.68 (7.55,12.42) & 3.77 (2.82,5.24) & 3.16 (2.56,3.63) & 6.68 (3.95,9.03) & 4.83 (3.10,7.19) & 6.20 (3.62,9.10) & 5.71 (3.65,7.94) \\
MSD-BrainTumour & Enhancing tumour & 13.54 (10.43,16.43) & 11.76 (7.16,15.23) & 10.36 (6.63,15.21) & 12.61 (8.50,17.21) & 7.44 (5.17,12.06) & 4.44 (3.09,6.46) & 2.22 (1.87,2.68) & 4.94 (3.39,7.76) & 3.85 (2.60,5.51) & 5.46 (3.30,7.98) & 4.79 (3.07,7.03) \\
MSD-BrainTumour & Non-enhancing tumor & 14.58 (10.72,20.55) & 11.91 (7.75,16.86) & 10.51 (6.92,15.60) & 13.91 (9.38,18.56) & 8.87 (5.92,13.71) & 4.44 (3.20,6.04) & 2.27 (1.97,2.84) & 7.17 (4.78,10.30) & 6.46 (4.64,9.58) & 7.17 (4.99,10.75) & 6.88 (4.73,10.53) \\
BraTS-FLAIR & Brain Tumor & 3.32 (2.54,4.55) & 3.50 (2.58,4.97) & 3.18 (2.23,4.36) & 3.26 (2.20,4.94) & 4.73 (3.54,6.01) & 2.36 (2.05,2.79) & 2.42 (2.08,2.91) & 2.24 (1.74,2.94) & 2.24 (1.70,3.02) & 1.93 (1.60,2.64) & 2.07 (1.65,3.07) \\
BraTS-T1 & Brain Tumor & 4.45 (3.52,5.82) & 4.64 (3.73,6.01) & 4.83 (3.60,6.04) & 5.23 (3.81,7.34) & 5.50 (4.44,7.11) & 4.13 (3.57,5.23) & 3.02 (2.60,3.45) & 3.29 (2.60,4.29) & 3.46 (2.58,4.32) & 2.88 (2.46,3.75) & 3.31 (2.49,4.02) \\
BraTS-T1CE & Brain Tumor & 2.32 (1.67,3.18) & 2.62 (1.68,3.27) & 2.31 (1.69,3.31) & 2.37 (1.78,3.29) & 2.48 (1.94,3.13) & 3.14 (2.70,3.73) & 1.90 (1.71,2.22) & 1.68 (1.32,2.46) & 1.55 (1.26,2.15) & 1.47 (1.25,2.40) & 1.67 (1.32,2.74) \\
ISLES2022-DWI & Ischemic Stroke & 3.91 (1.63,6.38) & 3.56 (2.20,6.30) & 3.22 (1.43,5.91) & 4.95 (1.86,6.41) & 1.53 (1.15,3.21) & 6.25 (4.78,7.63) & 1.17 (1.10,1.30) & 1.29 (1.05,1.83) & 1.32 (1.07,1.82) & 1.18 (1.03,1.67) & 1.29 (1.08,1.64) \\
ISLES2022-ADC & Ischemic Stroke & 3.25 (1.49,5.49) & 4.26 (1.99,6.80) & 2.92 (1.57,4.48) & 3.85 (2.05,6.62) & 2.82 (1.80,3.64) & 5.28 (4.13,6.66) & 1.22 (1.16,1.42) & 1.49 (1.17,2.33) & 1.60 (1.18,2.22) & 1.50 (1.19,2.52) & 1.42 (1.17,2.24) \\
\hline
\end{longtable}}
\end{landscape}

\begin{landscape}
\subsection*{H.\quad Statistical Significance}
\medskip\noindent{\footnotesize\textbf{Table S12.} Two-sided paired Wilcoxon signed-rank tests with SAMRI-3D as reference; rows grouped by metric and aggregation level.}\par\nobreak\smallskip

{\scriptsize\renewcommand{\arraystretch}{0.85}\setlength{\tabcolsep}{3pt}\setlength\LTleft{-0.85in}\setlength\LTright{-0.85in}
\begin{longtable}{|@{\extracolsep{\fill}}l|c|c|c|c|c|c|c|}
\hline
\textbf{Comparison (vs SAMRI-3D)} & \textbf{n} & \textbf{median (ours)} & \textbf{median (other)} & \textbf{median $\Delta$} & \textbf{win-rate (ours)} & \textbf{W} & \textbf{p-value} \\ \hline\endfirsthead
\hline \textbf{Comparison (vs SAMRI-3D)} & \textbf{n} & \textbf{median (ours)} & \textbf{median (other)} & \textbf{median $\Delta$} & \textbf{win-rate (ours)} & \textbf{W} & \textbf{p-value} \\ \hline\endhead
\multicolumn{8}{|l|}{\textbf{DSC --- Sample-level (n=3376)}} \\
SAMRI-3D-2dloss & 3376 & 0.746 & 0.734 & +0.0005 & 51.7\% & 2391185 & 9.75e-14 \\
GVT-v1 & 3376 & 0.746 & 0.740 & +0.0004 & 50.8\% & 2510184 & 3.42e-07 \\
GVT-v2 & 3376 & 0.746 & 0.749 & +0.0000 & 49.9\% & 2747288 & 2.17e-01 \\
SAMed-2 & 3376 & 0.746 & 0.642 & +0.0699 & 67.7\% & 1841746 & 6.47e-71 \\
SAM2-Small & 3376 & 0.746 & 0.523 & +0.1024 & 89.0\% & 306357 & 0.00e+00 \\
\hline
\multicolumn{8}{|l|}{\textbf{HD --- Sample-level (n=3376)}} \\
SAMRI-3D-2dloss & 3358 & 16.093 & 15.000 & +0.0000 & 43.8\% & 2210750 & 1.61e-02 \\
\hline
\end{longtable}}
\end{landscape}

\subsection*{I.\quad Additional Qualitative Examples}
Representative successful (Fig.~S2) and failure (Fig.~S3) cases across anatomical domains. Ground truth is shown in yellow and the SAMRI-3D prediction in pink.
\begin{center}
\includegraphics[height=0.78\textheight]{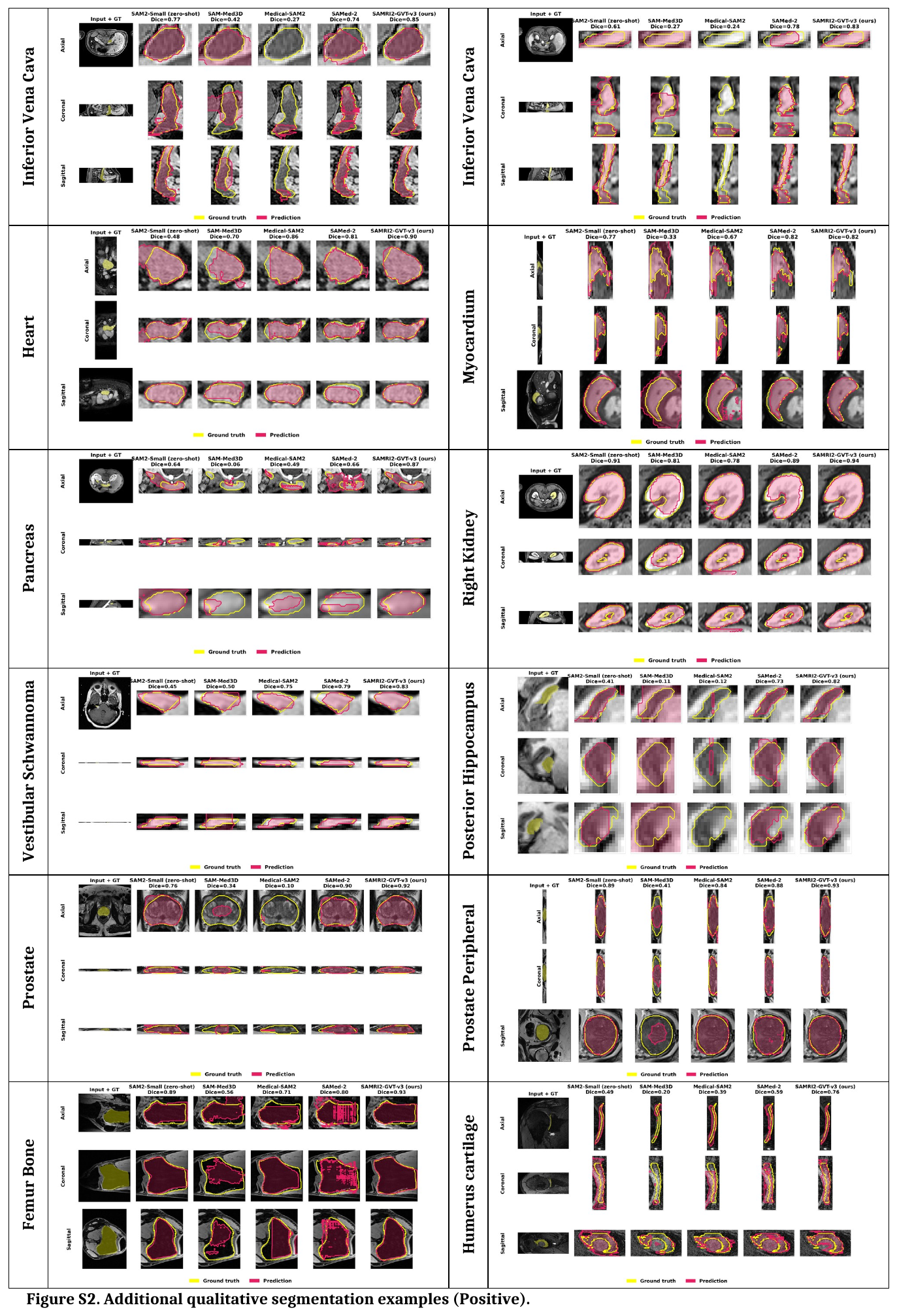}\\[3pt]
{\footnotesize\itshape \textbf{Figure S2.} Additional qualitative segmentation examples (successful cases).}
\end{center}

\clearpage
\begin{center}
\includegraphics[width=0.98\linewidth]{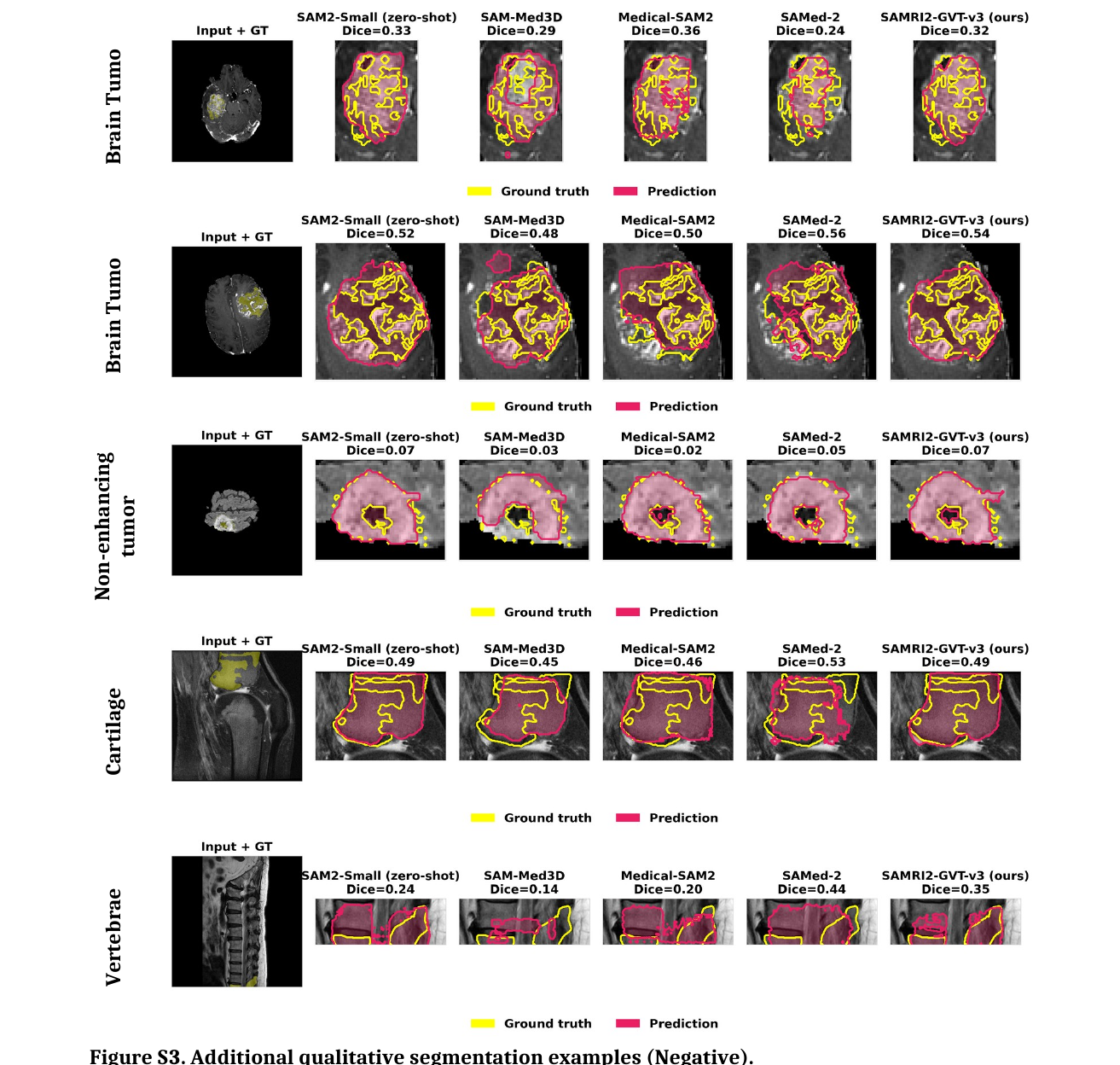}\\[3pt]
{\footnotesize\itshape \textbf{Figure S3.} Additional qualitative segmentation examples (failure cases).}
\end{center}

\subsection*{Dataset References}
{\footnotesize
\noindent [1] A. L. Simpson et al., ``A large annotated medical image dataset for the development and evaluation of segmentation algorithms,'' arXiv:1902.09063, 2019; M. Antonelli et al., ``The Medical Segmentation Decathlon,'' Nat. Commun., 13:4128, 2022.

\noindent [2] B. H. Menze et al., ``The multimodal brain tumor image segmentation benchmark (BRATS),'' IEEE Trans. Med. Imag., 34(10):1993--2024, 2015; S. Bakas et al., ``Advancing the cancer genome atlas glioma MRI collections with expert segmentation labels and radiomic features,'' Sci. Data, 4:170117, 2017; S. Bakas et al., ``Segmentation labels and radiomic features for the pre-operative scans of the TCGA-GBM and TCGA-LGG collections,'' The Cancer Imaging Archive, 2017; S. Bakas et al., ``Identifying the best machine learning algorithms for brain tumor segmentation\ldots,'' arXiv:1811.02629, 2018.

\noindent [3] M. R. Hernandez Petzsche et al., ``ISLES 2022: A multi-center MRI stroke lesion segmentation dataset,'' Sci. Data, 9:762, 2022.

\noindent [4] H. J. Kuijf et al., ``Standardized assessment of automatic segmentation of white matter hyperintensities and results of the WMH segmentation challenge,'' IEEE Trans. Med. Imag., 38(11):2556--2568, 2019.

\noindent [5] A. Saha, M. Hosseinzadeh, H. Huisman, ``End-to-end prostate cancer detection in bpMRI via 3D CNNs\ldots,'' Med. Image Anal., 73:102155, 2021 (PI-CAI).

\noindent [6] A. S. Becker et al., ``Variability of manual segmentation of the prostate in axial T2-weighted MRI: a multi-reader study,'' Eur. J. Radiol., 121:108716, 2019.

\noindent [7] Prostate (ADC) data from the CVPR 2024 challenge ``Segment Anything in Medical Images on Laptop'' (MedSAM on Laptop): J. Ma, Y. Zhou, B. Wang et al., ``Segment Anything in Medical Images and Videos: Benchmark and Deployment,'' arXiv:2408.03322, 2024; proceedings in Medical Image Segmentation Foundation Models (CVPR 2024 Challenge), LNCS 15458, Springer, 2025.

\noindent [8] A. Fedorov et al., ``An annotated test-retest collection of prostate multiparametric MRI,'' Sci. Data, 5:180281, 2018; hosted on The Cancer Imaging Archive: K. Clark et al., ``The Cancer Imaging Archive (TCIA): maintaining and operating a public information repository,'' J. Digit. Imaging, 26(6):1045--1057, 2013.

\noindent [9] G. Litjens et al., ``Evaluation of prostate segmentation algorithms for MRI: the PROMISE12 challenge,'' Med. Image Anal., 18(2):359--373, 2014.

\noindent [10] F. Ambellan et al., ``Automated segmentation of knee bone and cartilage combining statistical shape knowledge and CNNs: Data from the Osteoarthritis Initiative,'' Med. Image Anal., 52:109--118, 2019; source imaging from the Osteoarthritis Initiative (OAI): C. G. Peterfy, E. Schneider, M. Nevitt, ``The osteoarthritis initiative: report on the design rationale for the MRI protocol for the knee,'' Osteoarthritis Cartilage, 16(12):1433--1441, 2008.

\noindent [11] Osteoarthritis Initiative (OAI): C. G. Peterfy, E. Schneider, M. Nevitt, ``The osteoarthritis initiative: report on the design rationale for the MRI protocol for the knee,'' Osteoarthritis Cartilage, 16(12):1433--1441, 2008; https://nda.nih.gov/oai.

\noindent [12] O. Bernard et al., ``Deep learning techniques for automatic MRI cardiac multi-structures segmentation and diagnosis: is the problem solved?,'' IEEE Trans. Med. Imag., 37(11):2514--2525, 2018.

\noindent [13] D. F. Pace et al., ``HVSMR-2.0: A 3D cardiovascular MR dataset for whole-heart segmentation in congenital heart disease,'' Sci. Data, 11:721, 2024.

\noindent [14] T. A. D'Antonoli et al., ``TotalSegmentator MRI: robust sequence-independent segmentation of multiple anatomic structures in MRI,'' Radiology, 314(2):e241613, 2025.

\noindent [15] R. Dorent et al., ``CrossMoDA 2021 challenge: benchmark of cross-modality domain adaptation for vestibular schwannoma and cochlea segmentation,'' Med. Image Anal., 83:102628, 2023; original dataset: J. Shapey et al., ``Segmentation of vestibular schwannoma from MRI, an open annotated dataset and baseline algorithm,'' Sci. Data, 8:286, 2021.

\noindent [16] S. S. Chandra et al., ``Focused shape models for hip joint segmentation in 3D magnetic resonance images,'' Med. Image Anal., 18(3):567--578, 2014.

\noindent [17] D. Zukić et al., ``Robust detection and segmentation for diagnosis of vertebral diseases using routine MR images,'' Comput. Graph. Forum, 33(6):190--204, 2014.

\noindent [18] Y. Ji et al., ``AMOS: a large-scale abdominal multi-organ benchmark for versatile medical image segmentation,'' Adv. Neural Inf. Process. Syst., 35:36722--36732, 2022.

\noindent [19] S. R. Bowen et al., ``Tumor radiomic heterogeneity: multiparametric functional imaging to characterize variability and predict response following cervical cancer radiation therapy,'' J. Magn. Reson. Imaging, 47(5):1388--1396, 2018.
}

\end{document}